\documentclass[acmlarge]{acmart}
\AtBeginDocument{%
  }

\usepackage{multirow}
\usepackage{graphicx} 
\usepackage{subcaption} 
\usepackage{xcolor} 
\usepackage{caption}
\usepackage{tikz}
\usepackage{tcolorbox}
\usepackage{listings}
\tcbuselibrary{listings,breakable}
\usetikzlibrary{arrows.meta, positioning}

\setcopyright{none}
\makeatletter
\renewcommand{\@copyrightowner}{}
\renewcommand{\@copyrightpermission}{}
\renewcommand{\@formatdoi}[1]{}
\renewcommand{\@copyrightyear}{}
\makeatother


\makeatletter
\def\@acmReferenceFormat#1{} 
\def\@acmPages{} 
\makeatother

\begin{document}

\title{Infection-Reasoner: A Compact Vision-Language Model for Wound Infection Classification with Evidence-Grounded Clinical Reasoning}

\author{Palawat Busaranuvong}
\email{pbusaranuvong@wpi.edu}
\author{Reza Saadati Fard}
\email{rsaadatifard@wpi.edu}
\author{Emmanuel Agu}
\authornote{Corresponding author.}
\email{emmanuel@wpi.edu}
\author{Deepak Kumar}
\email{dkumar1@wpi.edu}
\author{Shefalika Gautam}
\email{sgautam@wpi.edu}
\author{Bengisu Tulu}
\email{bengisu@wpi.edu}
\author{Diane Strong}
\email{dstrong@wpi.edu}

\affiliation{%
  \institution{Worcester Polytechnic Institute}
  \city{Worcester}
  \state{MA}
  \country{USA}
}

\author{Giorgio Giatsidis}
\email{Giorgio.Giatsidis@umassmed.edu}

\affiliation{%
  \institution{UMass Chan Medical School}
  \city{Worcester}
  \state{MA}
  \country{USA}
}

\author{Lorraine Loretz}
\email{loretzdpmnp@gmail.com}

\affiliation{%
  \institution{UMass Memorial Healthcare (retired)}
  \city{Worcester}
  \state{MA}
  \country{USA}
}

\renewcommand{\shortauthors}{Busaranuvong et al.}

\begin{abstract}
Assessing chronic wound infection from photographs is challenging because visual appearance varies across wound etiologies, anatomical locations, and imaging conditions. Prior image-based deep learning methods have mainly focused on classification with limited interpretability, despite the need for evidence-grounded explanations to support point-of-care decision making. We present Infection-Reasoner, a compact 4B-parameter reasoning vision-language model for chronic wound infection classification and rationale generation. To address the scarcity of expert-labeled wound images with reasoning annotations, Infection-Reasoner is trained using a two-stage pipeline: (1) reasoning distillation, in which GPT-5.1 generates chain-of-thought rationales for unlabeled wound images to initialize wound-specific reasoning in a smaller student model (Qwen3-VL-4B-Thinking), and (2) reinforcement learning post-training with Group Relative Policy Optimization on a small labeled infection dataset to refine classification reasoning. On a held-out heterogeneous wound dataset, Infection-Reasoner achieved 86.8\% accuracy, 86.4\% sensitivity, and 87.1\% specificity, outperforming several strong baselines, including GPT-5.1. Rationale quality was further evaluated using both multimodal large language model (MLLM) judges and wound expert review. Across four MLLM judges, visual-support agreement scores ranged from 0.722 to 0.903, while expert review rated 61.8\% of rationales as Correct and 32.4\% as Partially Correct.

\end{abstract}



\settopmatter{printacmref=false}
\maketitle

\section{Introduction}
\label{sec:introduction}

Chronic wounds impose a substantial burden on patients, caregivers, and health systems. In the United States alone, millions of individuals are affected, with particularly high impact among older adults, and the aggregate economic burden exceeds \$10 billion annually across care settings~\cite{jarbrinkhumanistic, nussbaum2018economic, olsson2019humanistic, gould2015chronic}. Importantly, chronic wounds arise from diverse etiologies including diabetic foot ulcers, venous leg ulcers, arterial ulcers, pressure injuries, surgical wounds, and trauma-related wounds, each exhibiting different visual characteristics, healing trajectories, and clinical contexts~\cite{bowers2020chronic, fonder2008treating, franks2016management}. This heterogeneity makes automated wound assessment particularly challenging in real-world settings.

Among wound complications, infection is one of the most clinically consequential because it can rapidly destabilize a wound that was previously manageable~\cite{richard2011new, lipsky2012idf}. Wound infection typically involves microbial invasion of host tissue followed by inflammatory responses that disrupt normal healing~\cite{landis2008chronic, mcdonald2018infectious}. Clinically, infection diagnosis relies on interpretation of a constellation of signs and symptoms rather than a single visual cue. Common indicators include erythema, edema, warmth, tenderness or pain, purulent exudate, tissue friability, necrosis, worsening wound appearance, and delayed healing~\cite{lipsky2012idf, iwgdf2023dfi, stallard2018and}. In chronic wounds, these signs can be subtle or atypical, complicating early recognition and increasing the risk of delayed treatment, hospitalization, or limb-threatening complications and amputations~\cite{mills2014society}.

In routine clinical practice, diagnosing infected wounds often involves multiple steps including wound debridement, microbiological cultures, laboratory tests, and specialist interpretation. While appropriate in well-resourced clinical environments, this workflow can be difficult to execute at the point-of-care (POC), such as during home visits, outpatient follow-up, or in resource-limited settings~\cite{lipsky2016antimicrobial, maclellan2002designing}. In these contexts, non-specialist caregivers frequently lack immediate access to wound experts and instead make assessments based on wound appearance and bedside observations. Consequently, triage decisions must be made under uncertainty: infected wounds may be referred too late, while many non-infected wounds are unnecessarily escalated, increasing healthcare cost and workflow burden~\cite{rondas2015prevalence, wilbright2004use, chanussot2013telemedicine}.

Although wound photography is increasingly integrated into telehealth, reliable infection assessment from images alone remains challenging. Real-world wound photographs vary widely in lighting conditions, viewpoint, focus, resolution, background clutter, and anatomical location. In addition, visual phenotypes often overlap between inflammation, colonization, healing tissue, and true infection~\cite{scebba2021detectsegment, shenoy2018deepwound}. These factors create substantial intra-class variability and inter-class similarity, reducing confidence in appearance-based classification. Prior deep learning approaches have demonstrated promising results, but most systems are developed primarily on datasets containing diabetic foot ulcers (DFUs)~\cite{partb_DFU, yap2021analysis} and often rely on explicit wound localization or cropped patches prior to classification~\cite{wang2015unified, al2022diabetic, ConDiff, qayyum2021vit, galdran2021convolutional}. Such assumptions may limit generalization when wound types and acquisition conditions vary across clinical environments. Beyond generalization limitations, most existing deep learning systems operate as black-box classifiers that output only a prediction without clinically meaningful explanations. In medical decision support settings, such opaque predictions can limit clinician trust and make it difficult to verify whether the model’s decision is grounded in evidence-based wound signs. The absence of structured clinical reasoning or textual explanations of machine learning assessments therefore remains a major barrier to deploying automated wound assessment systems in real-world care workflows.

Meanwhile, Multimodal Large Language Models (MLLMs) have rapidly advanced. Large vision-language models (VLMs) such as GPT-5.1~\cite{gpt5}, Claude-4.6~\cite{claude4.6}, and Gemini-3~\cite{gemini3pro} have demonstrated strong image--text reasoning capabilities and have begun to show promising performance on medical reasoning benchmarks~\cite{med-gemini, safavi2025gastrobenchmark, kaczmarczyk2024evaluating}. 
However, these MLLMs typically require substantial GPU memory, high inference cost, and cloud-scale computational infrastructure, which can limit their practicality for routine deployment in resource-constrained clinical settings. In parallel, smaller open-source medical VLMs (roughly 2B--32B parameters) including MedGemma~\cite{google2025medgemma}, Lingshu~\cite{xu2025lingshu}, and MedVLM-R1~\cite{pan2025medvlmR1} suggest that compact models can support clinically meaningful decision-making with performance comparable to larger MLLMs on task-specific problems but with lower inference cost, making deployment on edge or mobile devices more feasible. However, current medical VLM development has not explored wound assessment from photographs, focusing instead on structured imaging modalities such as X-ray, CT, MRI, pathology slides, or fundus imaging~\cite{tu2024towards, zhang2023huatuogpt}. This gap, together with the practical need for low-cost and interpretable decision support at the POC, motivates our development of a compact wound-specific reasoning vision-language model.

We propose \textbf{Infection-Reasoner}, a 4B-parameter reasoning VLM designed for end-to-end chronic wound infection classification and clinical rationale generation from wound photographs. Unlike prior DFU-specific frameworks, Infection-Reasoner directly analyzes whole wound images across heterogeneous wound types without requiring explicit image preprocessing such as wound localization. The model is trained using a two-stage pipeline:
(1) \textit{reasoning distillation}, in which a large teacher model (GPT-5.1) generates pseudo-label predictions and corresponding chain-of-thought (CoT) rationales for unlabeled wound images, which are then used to initialize cold-start reasoning via supervised fine-tuning (SFT) of the student model.
This stage addresses the scarcity of expert-written diagnostic rationales by transferring structured wound-sign reasoning without manual rationale annotation; and (2) \textit{reinforcement-learning (RL) post-training} on limited labeled wound infection data to further improve infection classification accuracy and the quality of clinical rationale generated. 
Since distillation alone can inherit teacher bias and may generalize less robustly under domain shift, especially when the teacher is not wound-specialized, combining the second RL stage provides task-specific adaptation to the wound infection task.

Consequently, Infection-Reasoner-4B achieves strong performance on an unseen heterogeneous wound dataset, reaching 86.8\% accuracy, 86.4\% sensitivity, and 87.1\% specificity, outperforming its ablated, individual training components as well as proprietary foundation models such as GPT-5.1, Gemini-3.1, and Claude-4.6, evaluated in a CoT-prompted zero-shot setting. Beyond infection classification, Infection-Reasoner produces wound-sign-aware clinical rationales that align visible evidence with diagnostic reasoning, supporting interpretable decision-making for nurses and clinicians at the point of care. At the same time, because our labeled training and evaluation sets are limited in size, we position these results as a first, important step that provides preliminary evidence of the potential of post-trained, compact wound-specific reasoning models rather than as definitive clinical validation.

\textbf{Our main contributions} are as follows:

\begin{itemize}

\item We introduce \textbf{Infection-Reasoner}, a compact reasoning VLM for end-to-end infection classification and rationale generation from whole wound images without explicit image preprocessing steps, such as, wound localization, segmentation, and cropping.


\item We propose a \textbf{two-stage training pipeline}: (1) reasoning distillation on teacher-generated CoT data via SFT, followed by (2) RL post-training on limited labeled data. This design is motivated by a key practical constraint in wound-care AI: expert-labeled infection images with corresponding image-grounded rationale annotations are both scarce. Reasoning distillation allows the model to learn structured wound-sign reasoning from teacher generated rationales, while RL post-training reduces residual teacher bias and sharpen decision boundaries while preserving the reasoning structure learned in the first stage.

\item Extensive experiments demonstrate that \textbf{our Infection-Reasoner model achieves strong diagnostic performance} on unseen heterogeneous chronic wound images (DFU, pressure, venous, arterial), outperforming MLLMs evaluated with CoT-prompted zero-shot settings. 
These results demonstrate that the proposed 2-stage Infection-Reasoner framework enables a compact 4B model to surpass proprietary MLLMs, including GPT-5.1, on a specialized wound infection imaging task.

\item We introduce an \textbf{MLLM-as-a-Judge evaluation protocol} to assess rationale grounding. This protocol extracts clinical claims from model-generated reasoning and verifies whether those claims are supported by visual evidence in the wound image. 
We further complement the reasoning evaluation with expert review, providing human validation of rationale plausibility and clinical usefulness.

\end{itemize}

\section{Related Work}
\label{sec:RelatedWork}

\subsection{Deep Learning for Wound Infection Classification}

\begin{table}[ht]
\centering
\caption{Summary of prior work on wound infection classification using deep learning.
}
\resizebox{\textwidth}{!}{
\begin{tabular}{cccccc}
\hline
\textbf{Task} &
  \textbf{Related Work} &
  \textbf{\begin{tabular}[c]{@{}c@{}}Summary of \\ Approach\end{tabular}} &
  \textbf{\begin{tabular}[c]{@{}c@{}}No. of \\ Target Classes\end{tabular}} &
  \textbf{Dataset} &
  \textbf{Results} \\ \hline
\centering
\begin{tabular}[c]{@{}c@{}}Wound segmentation \\ and Infection \\ Classification\end{tabular} &
  \begin{tabular}[c]{@{}c@{}}Wang et al. \\ 2015 \cite{wang2015unified}\end{tabular} &
  \begin{tabular}[c]{@{}c@{}}CNN-based: \\ ConvNet + SVM\end{tabular} &
  \begin{tabular}[c]{@{}c@{}}2 classes \\ (infection and\\  no infection)\end{tabular} &
  \begin{tabular}[c]{@{}c@{}}NYU wound \\ Database\end{tabular} &
  \begin{tabular}[c]{@{}c@{}}Accuracy: 95.6\%\\ PPV: 40\%\\ Sensitivity: 31\%\end{tabular} \\ \hline 
\centering
 &
  \begin{tabular}[c]{@{}c@{}}Goyal et al.\\ 2020 \cite{partb_DFU} \end{tabular} &
  \begin{tabular}[c]{@{}c@{}}CNN-based: \\ Ensemble CNN\end{tabular} &
  \multirow{3}{*}{\begin{tabular}[c]{@{}c@{}} \\ \\ 2 classes \\ (infection and \\ no infection)\end{tabular}} &
  \multirow{3}{*}{\begin{tabular}[c]{@{}c@{}} \\ \\  Part B DFU \\  2020 dataset\\ \end{tabular}} &
  \begin{tabular}[c]{@{}c@{}}Accuracy: 72.7\%\\ PPV: 73.5\%\\ Sensitivity: 70.9\%\end{tabular} \\ \cline{2-3} \cline{6-6} 
\centering
\multirow{-2}{*}{\begin{tabular}[c]{@{}c@{}}DFU infection \\ classification\end{tabular}} &
  \begin{tabular}[c]{@{}c@{}}Al-Garaawi et al. \\ 2022 \cite{al2022diabetic} \end{tabular} &
  \begin{tabular}[c]{@{}c@{}}CNN-based:\\ DFU-RGB-TEX-Net\end{tabular} &
   &
   &
  \begin{tabular}[c]{@{}c@{}}Accuracy: 74.2\%\\ PPV: 74.1\%\\ Sensitivity: 75.1\%\end{tabular} \\ \cline{2-3} \cline{6-6} 
\centering
 &
   \begin{tabular}[c]{@{}c@{}}Busaranuvong et al. \\ 2025 \cite{scarwid} \end{tabular} &
  \begin{tabular}[c]{@{}c@{}} Multimodal Transformer:\\ SCARWID (retrieval-based)\end{tabular} &
   &
   &
  \begin{tabular}[c]{@{}c@{}} Accuracy: 81.4\%\\ PPV: 79.4\%\\ Sensitivity: 84.5\%\end{tabular} \\ \hline
\centering
\multirow{3}{*}{\begin{tabular}[c]{@{}c@{}}\\ \\ DFU wound ischemia\\ and  infection \\ classification\end{tabular}} &
  \begin{tabular}[c]{@{}c@{}}Yap et al. \\ 2021 \cite{yap2021analysis} \end{tabular} &
  \begin{tabular}[c]{@{}c@{}}CNN-based:\\ InceptionV3, DenseNet, \\ EfficientNet\end{tabular} &
  \multirow{3}{*}{\begin{tabular}[c]{@{}c@{}}\\ 4 classes\\ (both infection\\ and ischemia,\\ infection, ischemia,\\ none)\end{tabular}} &
  \multirow{3}{*}{\begin{tabular}[c]{@{}c@{}} \\ \\  DFUC2021\\ dataset\end{tabular}} &
  \begin{tabular}[c]{@{}c@{}}EfficientNet B0 \\ performance: \\ F1, PPV, SEN\\ =  55\% , 57\%, 62\%\end{tabular} \\ \cline{2-3} \cline{6-6} 
\centering
 &
  \begin{tabular}[c]{@{}c@{}}Qayyum et al. \\ 2021 \cite{qayyum2021vit} \end{tabular} &
  \begin{tabular}[c]{@{}c@{}}ViT-based: \\
  Ensemble ViT \end{tabular} &
   &
   &
  \begin{tabular}[c]{@{}c@{}} F1, PPV, SEN\\ = 57\%, 58\% , 61\%\end{tabular} \\ \cline{2-3} \cline{6-6} 
\centering
 &
  \begin{tabular}[c]{@{}c@{}}Galdran et al. \\ 2021 \cite{galdran2021convolutional} \end{tabular} &
  \begin{tabular}[c]{@{}c@{}}ViT-based: ViT, DeiT 
  \\ CNN-based: BiT, \\ EfficientNet\end{tabular} &
   &
   &
  \begin{tabular}[c]{@{}c@{}}BiT performance:\\ F1, PPV, SEN\\ = 61\%, 61\% , 66\%\end{tabular} \\ \hline
\end{tabular}
}
\label{tab:summary_of_approaches}
\end{table}

State-of-the-art (SOTA) deep learning wound infection detection approaches have primarily focused on image-based classification using Convolutional Neural Networks (CNNs) and vision transformers (ViTs)~\cite{partb_DFU, galdran2021convolutional, yap2021analysis, al2022diabetic, qayyum2021vit}. As summarized in Table~\ref{tab:summary_of_approaches}, the large majority of such research has been conducted on datasets containing diabetic foot ulcer (DFU) images, which represent only one subtype of chronic wounds. 
Other major chronic wound subtypes include venous ulcers, arterial ulcers, and pressure injuries~\cite{bowers2020chronic}. Because these wound types differ in pathophysiology, visual phenotype, and clinically relevant signs for infection assessment~\cite{fonder2008treating},
models trained primarily on DFU images may not generalize reliably across heterogeneous chronic wounds.

Early work by Wang et al.~\cite{wang2015unified} utilized Support Vector Machines (SVM) to classify image features extracted by a CNN, achieving high overall accuracy. However, the low sensitivity indicated the difficulty of correctly identifying infected cases. Later DFU-focused systems by Goyal et al.~\cite{partb_DFU}, Al-Garaawi et al.~\cite{al2022diabetic}, Yap et al.~\cite{yap2021analysis}, Qayyum et al.~\cite{qayyum2021vit}, and Galdran et al.~\cite{galdran2021convolutional} improved classification performance using ensemble CNNs, texture-augmented networks, and transformer-based architectures. However, these methods are primarily predictive rather than explanatory: they output a class label without generating clinically meaningful textual rationales, limiting interpretability and clinician trust. These approaches also depend on explicit wound localization prior to classification, which may reduce robustness when acquisition conditions vary across real-world settings.

Recent work has begun to incorporate language or multimodal representations into wound analysis. Synthetic Caption Augmented Retrieval for Wound Infection Detection (SCARWID)~\cite{scarwid}, for example, augments wound images with synthetic captions generated by a fine-tuned VLM to improve DFU infection classification. It moves beyond pure image classification by introducing text-based auxiliary information, but the captions generated mainly serve as feature augmentation for the downstream classifier rather than as complete, standalone diagnostic explanations or explicit clinical rationale. 

Across all prior systems, two limitations persist. First, most models are developed on relatively narrow wound distributions, especially DFU-focused datasets, which may limit generalization across heterogeneous chronic wounds. Second, existing models produce only a classification label (e.g. infected or uninfected) without any corresponding clinically grounded textual rationale, limiting interpretability and clinician trust in POC deployment. These gaps motivate the development of reasoning-capable VLMs that can both classify infection and generate clinically grounded rationales directly from wound images.


\subsection{Medical Vision-Language Models}

The rapid development of large vision-language models (VLMs) has opened new possibilities for multimodal clinical decision support. Early work in this area centered on medical visual question answering (Med-VQA), with evaluation on benchmarks such as SLAKE~\cite{liu2021slake}, VQA-RAD~\cite{lau2018vqavad}, and PathVQA~\cite{he2020pathvqa}. These benchmarks helped driving the development of models capable of interpreting radiology, pathology, and other biomedical images through image--text reasoning.

Large foundation models have demonstrated strong performance on broader medical reasoning tasks. Med-PaLM Multimodal~\cite{tu2024towards}, for example, showed that scaling a pretrained vision encoder with a large language model backbone enables competitive performance across diverse medical imaging benchmarks by achieving F1-scores of 62.06\% on VQA-RAD, 62.69\% on Path-VQA, and 87.50\% on SLAKE-VQA. HuatuoGPT-Vision~\cite{zhang2023huatuogpt} further showed that domain-aligned instruction tuning benefits medical VQA and report generation, while MedGemma~\cite{google2025medgemma} illustrated that compact open-weight models can still support clinically meaningful decision-making when fine-tuned on curated biomedical corpora. Proprietary general-purpose VLM families such as GPT~\cite{gpt4, gpt5}, Gemini~\cite{comanici2025gemini, gemini3pro}, and Claude~\cite{claude3, claude4.6} have also been evaluated on medical reasoning benchmarks~\cite{med-gemini, safavi2025gastrobenchmark, kaczmarczyk2024evaluating}, demonstrating strong CoT-prompted zero-shot capabilities, but typically with higher inference latency, greater memory requirements, and higher computational cost than compact specialized models.

 Despite this progress, most medical VLM development has focused on relatively structured imaging domains such as radiology, pathology, and fundus photography~\cite{tu2024towards, med-gemini, zhang2023huatuogpt, google2025medgemma}. In these settings, image acquisition is typically more standardized and large annotated datasets are more readily available. Wound photographs, by contrast, exhibit substantially greater variability in appearance, acquisition conditions, and clinical context, while annotated wound datasets remain limited. Recent reasoning-oriented models such as Med-R1~\cite{lai2026med}, MedVLM-R1~\cite{pan2025medvlmR1}, and Lingshu~\cite{xu2025lingshu} suggest that explicit reasoning supervision can improve performance and interpretability in medical multimodal systems, but they have been evaluated primarily on medical VQA-style benchmarks rather than wound assessment from real-world photographs. This leaves an important gap in compact, reasoning-capable models for end-to-end wound infection classification and evidence-grounded rationale generation at the POC.

\subsection{Reasoning Distillation and Reinforcement Learning for Medical AI}

Chain-of-Thought (CoT) prompting~\cite{wei2022chain} demonstrated that eliciting intermediate reasoning steps before a final prediction substantially improves LLM performance on complex inference tasks. In medical settings, this is particularly valuable because diagnosis often requires combined analyses of multiple visual and clinical cues rather than any single observation~\cite{jin2024hidden}. The success of DeepSeek-R1~\cite{guo2025deepseek} showed that training models to produce long-form reasoning traces via reinforcement learning can yield large performance gains on various tasks including
mathematical reasoning, programming, and other complex inference benchmarks that reward multi-step deliberation and verifiable answer correctness, catalyzing a wave of reasoning-oriented adaptations in the medical domain~\cite{lai2026med, pan2025medvlmR1, zhang2025med}.

A core challenge in training medical reasoning models is to obtain high-quality rationales at scale. Expert-written reasoning traces are tedious, expensive and challenging to collect. Wound photographs in particular rarely have corresponding structured diagnostic explanations. \textit{Reasoning distillation}, where a larger and more powerful teacher model generates CoT-style supervision to initialize a smaller student model, offers a practical alternative to manual annotation of rationale. Prior work has shown that teacher-generated reasoning can transfer useful step-by-step diagnostic patterns to compact student models effectively~\cite{li2023symbolic, bercovich2025llama, guo2025deepseek}. However, distillation alone can inherit teacher biases and may generalize less robustly under domain shift, particularly when the teacher was not trained on wound-specific data~\cite{li2023llavamed}.

\textit{Reinforcement Learning with Verifiable Rewards} (RLVR)~\cite{shao2024deepseekmath, guo2025deepseek} is a complementary approach that replaces learned reward models with automatically verifiable signals such as answer correctness. Group Relative Policy Optimization (GRPO)~\cite{shao2024deepseekmath} has emerged as an effective RLVR algorithm that optimizes model outputs relative to a group of sampled responses without requiring a separate critic model, making it practical under limited labeled data. Compared with Reinforcement Learning from Human Feedback (RLHF)~\cite{ouyang2022training}, which relies on human preference annotations to optimize model behavior, RLVR is substantially more feasible when such annotations are unavailable. However, applying RLVR alone can produce incoherent reasoning chains that lack human-readable structure~\cite{guo2025deepseek} and may not strengthen the intrinsic reasoning priors of the base model~\cite{yue2025does}.

These observations motivated us to apply the two-stage training framework, where \textit{Stage 1} applies reasoning distillation from a large teacher model, establishing a coherent wound-sign vocabulary and reasoning structure in the student model. \textit{Stage 2} then applies GRPO post-training to  sharpen infection classification, and refine rationale quality while preserving the human-readable explanation structure instilled in Stage 1.


\section{Infection-Reasoner}
\label{sec:infection_reasoner}

\begin{figure}[!th]
\centering
\includegraphics[width=0.9 \linewidth]{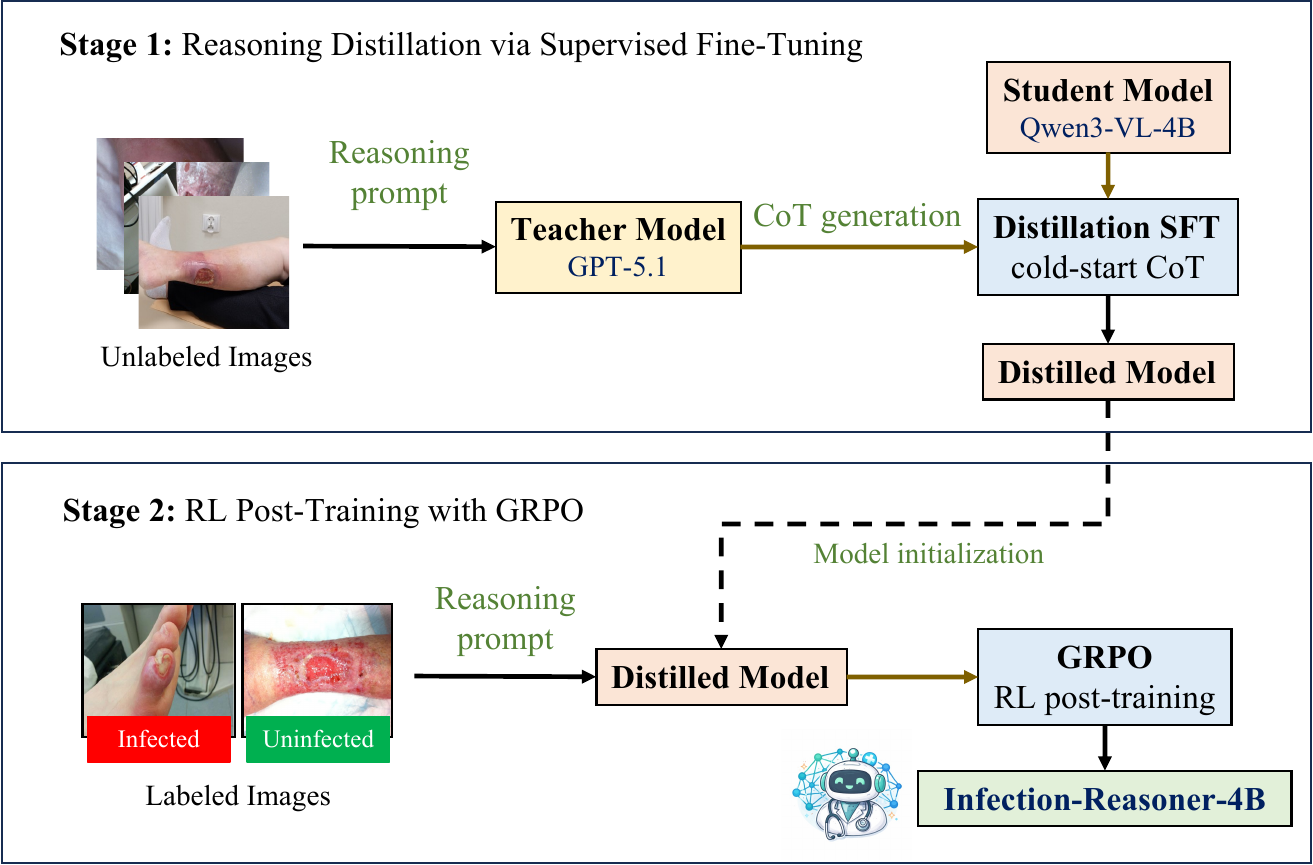}
\caption{The Infection-Reasoner Two-stage Training Pipeline. 
} 
\label{fig:framework}
\end{figure}

We instantiate Infection-Reasoner with Qwen3-VL-4B-Thinking~\cite{qwen3vl} as the base student model. We selected this model for its native dynamic-resolution processing, which allows whole wound photographs to be analyzed at their original resolution without fixed-size resizing, wound segmentation, or cropping. Preserving the full image in this way helps retain fine-grained visual cues relevant to infection assessment, such as exudate texture, perilesional skin color, and wound-edge morphology. Rather than modifying the model architecture, we adapt the base model to wound-specific reasoning through a two-stage training pipeline.

This design is motivated by the availability of wound photographs under two distinct supervision regimes. Unlabeled wound images can be curated at relatively larger scale from internet sources, whereas expert-labeled infection datasets are typically small and rarely include structured clinical reasoning annotations. Infection-Reasoner is designed to leverage both regimes through a two-stage training pipeline inspired by DeepSeek-R1~\cite{guo2025deepseek}, which showed that a cold-start supervised fine-tuning (SFT) phase before RL post-training can improve the coherence and usefulness of model reasoning. In Stage 1, a teacher model (GPT-5.1) generates chain-of-thought rationales for unlabeled wound images, and these rationales are distilled into the student model through SFT. In Stage 2, the distilled model is further optimized using GRPO~\cite{shao2024deepseekmath} on a small labeled infection dataset to refine classification behavior, reduce residual teacher bias, and improve performance under limited supervision. The overall Infection-Reasoner pipeline is illustrated in Figure~\ref{fig:framework}, and each stage is described in detail below.

\subsection{Stage 1: Reasoning Distillation Fine-Tuning (DFT)}
\label{sec:stage1}

Compact models trained with standard supervised learning on classification labels alone do not acquire structured diagnostic reasoning. Stage 1 addresses this challenge by using a large teacher model to generate wound-specific CoT rationales for unlabeled wound images, which are then distilled into the 4B-parameter student model.

\subsubsection*{Teacher-Generated Reasoning}

Given a curated set of unlabeled wound photographs $\mathcal{D}_u = \{x_i\}_{i=1}^{N}$, the teacher model is prompted to generate a CoT rationale followed by a binary infection classification for each image. The prompt in Figure~\ref{fig:reasoning_prompt} poses the task as a multiple-choice question (A: No / B: Yes) and constrains the model to analyze visual evidence only, explicitly excluding clinical history, laboratory results, odor, and systemic signs. Crucially, the prompt does \textit{not} enumerate specific wound signs to examine. This design allows the teacher's wound-sign vocabulary and diagnostic reasoning to emerge from its own clinical knowledge rather than being injected by the prompt. 
The result is a distillation dataset $\mathcal{D}_{\text{distill}} = \{(x_i, r_i, \hat{y}_i)\}_{i=1}^{N}$ of image--rationale--prediction triples, where $r_i$ is the teacher-generated CoT reasoning enclosed in \texttt{<think>}...\texttt{</think>} tags and $\hat{y}_i \in \{$\texttt{A}$,$\texttt{B}$\}$ is the teacher's answer enclosed in \texttt{<answer>}...\texttt{</answer>} tags.

\begin{figure}[!h]
\centering
\begin{tcolorbox}[
  colback=gray!5,
  colframe=gray!40,
  boxrule=0.5pt,
  arc=4pt,
  left=8pt, right=8pt, top=6pt, bottom=6pt,
  title={\small \textbf{Chain-of-Thought Reasoning Prompt (Stage 1 and Stage 2)}},
  fonttitle=\small,
  coltitle=black,
]
\small
Based on visual appearance, is the wound in the user-provided image infected?\\[4pt]
\textbf{Choice:} (A) No.\quad (B) Yes.\\[6pt]
First output the decision-making rationale in \texttt{<think></think>} tags and then output ONLY the final answer (i.e.\ single-letter choice) in \texttt{<answer></answer>} tags.\\[4pt]
Please strictly follow the format.\\[2pt]
\textbf{Formatting:} \texttt{<think>}...\texttt{</think>} \texttt{<answer>}...\texttt{</answer>}\\[4pt]
\textbf{Constraints:}
\begin{itemize}
  \item Use only visual evidence from the image; no clinical history, labs, odor, or systemic signs.
\end{itemize}
\end{tcolorbox}
\caption{Reasoning prompt used for teacher and student CoT generation in Stage 1 and for student RL post-training in Stage 2.}
\label{fig:reasoning_prompt}
\end{figure}

\subsubsection*{Supervised Fine-Tuning Objective}

The student model $\pi_\theta$ is fine-tuned on $\mathcal{D}_{\text{distill}}$ by maximizing the likelihood of the teacher-generated rationale and prediction given the input image. The training objective is the standard autoregressive language modeling loss~\cite{bengio2003neural, guo2025deepseek} over the target sequence $s_i = (r_i, \hat{y}_i)$:


\begin{equation}
\mathcal{L}_{\text{SFT}}(\theta) = -\mathbb{E}_{(x_i, s_i)\sim \mathcal{D}_{\text{distill}}}\left[\sum_{t=1}^{|s_i|} \log \pi_\theta\!\left(s_i^{(t)} \;\middle|\; x_i,\, s_i^{(<t)}\right)\right],
\label{eq:sft_loss}
\end{equation}

\noindent where $s_i^{(t)}$ denotes the $t$-th token of the target sequence and $s_i^{(<t)}$ denotes all preceding tokens. This loss function encourages the student to internalize the full diagnostic reasoning chain rather than learning only to predict the label.



\subsection{Stage 2: RL Post-Training with GRPO}
\label{sec:stage2}

To further align the distilled model with verified infection labels and refine reasoning, Stage 2 applies Group Relative Policy Optimization (GRPO)~\cite{shao2024deepseekmath} on a small labeled infection dataset, $\mathcal{D}_{\text{Labeled}} = \{(x_i, y_i)\}_{i=1}^{M}$, where $y_i \in \{\texttt{A}, \texttt{B}\}$ denotes the verified infection label for image $x_i$ (\texttt{A} = uninfected, \texttt{B} = infected), and $M$ is the number of labeled images. In this setting, GRPO refines the distilled model using direct supervision from verified infection outcomes. This design is further motivated by recent evidence that RL post-training for reasoning can remain effective even under very limited training data~\cite{wang2025reinforcement}.


\subsubsection*{Group Relative Policy Optimization} GRPO is a RL algorithm designed for language model post-training that eliminates the need for a separate critic or value network. For each input $x_i$, GRPO samples a group of $G$ candidate outputs $\{o_i^{(1)}, \ldots, o_i^{(G)}\}$ from the current policy $\pi_\theta$ and computes a reward $\mathcal{R}(o_i^{(g)}, y_i)$ for each. The relative advantage of each output within the group is estimated by normalizing the rewards:

\begin{equation}
\hat{A}_i^{(g)} = \frac{\mathcal{R}(o_i^{(g)}, y_i) - \text{mean}_g\bigl(\mathcal{R}(o_i^{(g)}, y_i)\bigr)}{\text{std}_g\bigl(\mathcal{R}(o_i^{(g)}, y_i)\bigr)},
\label{eq:grpo_advantage}
\end{equation}

\noindent where the mean and standard deviation are computed over the $G$ sampled outputs for the same input. The policy is then updated to maximize the clipped surrogate objective with a KL divergence penalty against a reference policy $\pi_{\text{ref}}$ (the Stage 1 distilled model):


\begin{equation}
\mathcal{L}_{\text{GRPO}}(\theta) =-\mathbb{E}_{\substack{
(x_i, y_i)\sim \mathcal{D}_{\text{Labeled}} \\
\{o_i^{(g)}\}_{g=1}^{G} \sim \pi_{\theta_{\text{old}}}(\cdot \mid x_i)}}
\left[\frac{1}{G}\sum_{g=1}^{G} \min\!\left(\rho_i^{(g)}\hat{A}_i^{(g)},\;
\text{clip}\!\left(\rho_i^{(g)}, 1-\varepsilon, 1+\varepsilon\right)\hat{A}_i^{(g)}
\right) -\beta\, \mathbb{D}_{\text{KL}}\!\left[\pi_\theta \,\|\, \pi_{\text{ref}}\right] \right]
\label{eq:grpo_loss}
\end{equation}

\noindent where $\rho_i^{(g)} = \pi_\theta(o_i^{(g)} | x_i) / \pi_{\theta_{\text{old}}}(o_i^{(g)} | x_i)$ is the importance sampling ratio, $\varepsilon$ is the clipping threshold, and $\beta$ controls the strength of the KL penalty. The KL term anchors the updated policy to the distilled model, preventing catastrophic forgetting of the reasoning structure established in Stage 1.

\subsubsection*{Accuracy Reward Function}

We define a single verifiable accuracy reward based on whether the model's predicted answer matches the ground-truth label $y_i$:

\begin{equation}
\mathcal{R}_{\text{acc}}(o_i^{(g)}, y_i) =
\begin{cases}
1 & \text{if } \texttt{parse\_answer}(o_i^{(g)}) = y_i, \\
0 & \text{otherwise,}
\end{cases}
\label{eq:reward}
\end{equation}

\noindent where $\texttt{parse\_answer}(\cdot)$ extracts the single-letter response (\texttt{A} or \texttt{B}) from within the \texttt{<answer>}...\texttt{</answer>} tags of the model output. If the model fails to produce a parseable answer in the required format, a reward of 0 is assigned. This binary verifiable reward provides a clean, unambiguous training signal that directly optimizes infection classification accuracy without requiring expert-annotated rationales~\cite{shao2024deepseekmath, guo2025deepseek}.

 In prior work that applies GRPO directly to a base model without cold-start initialization, a format reward is typically necessary to encourage the model to produce structured reasoning outputs~\cite{guo2025deepseek}. In this work, we do not include a format reward in Stage 2 since the structured \texttt{<think>}...\texttt{</think>} format and wound-sign reasoning vocabulary have already been transferred from the teacher in Stage 1. Adding a format reward on top of an already well-structured model would provide no additional training signal and could interfere with the nuanced reasoning patterns established during distillation. 

\section{Dataset Preparation and Processing}
\label{sec:dataset}

Infection-Reasoner was trained and evaluated using two distinct data regimes that correspond directly to its two training stages: a large pool of unlabeled wound images for Stage 1 distillation, and a smaller labeled infection dataset for Stage 2 RL post-training. Both datasets are drawn from heterogeneous sources to support generalization across wound types and acquisition conditions. 

In collaboration with the UMass Chan Medical School team (UMass), we acquired 120  wound images with unique IDs, collected during routine care photographs across 4 major chronic wound types: diabetic foot, pressure, venous, and arterial  ulcers. The images were collected with varying lighting conditions, viewpoints, and anatomical locations. The dataset collected by UMass is imbalanced, comprising 92 uninfected and 28 infected wounds, reflecting the lower prevalence of infection in outpatient wound care settings. Many severe infections are first managed through emergency services before referral.

\subsubsection*{Held-Out Test Set}
To obtain a clinically meaningful and fair evaluation protocol, we constructed a balanced held-out UMass test set by using all 28 infected wounds and randomly sampling 28 uninfected wounds (56 total). To prevent data leakage, all test images were excluded from both training stages.


\subsection{Teacher-Generated CoT Distillation Dataset}
\label{sec:gen_data}

Stage 1 was designed to learn wound-sign reasoning from a heterogeneous pool of \emph{unlabeled} wound images, without requiring infection labels or expert-written rationales. We curated 155 unique wound images from five sources: 33 from WoundNet~\cite{liu2021comprehensive}, 50 from the Medetec wound database~\cite{MedetecWebsite}, 16 from a clinical wound assessment report~\cite{WUWHS2016Triangle}, 10 from Internet image search, and 46 uninfected UMass wound images not included in the held-out test split. Table~\ref{tab:stage1_sources} summarizes the source dataset composition. 

\begin{table}[!h]
\centering
\caption{Composition of Stage 1 source for unlabeled distillation images.}
\label{tab:stage1_sources}
\begin{tabular}{lcc}
\hline
\textbf{Source} & \textbf{Count} & \textbf{Label usage} \\
\hline
WoundNet~\cite{liu2021comprehensive} & 33 & Not utilized \\
Medetec database~\cite{MedetecWebsite} & 50 & Not utilized\\
Wound Assessment report~\cite{WUWHS2016Triangle} & 16 & Not utilized\\
Internet image search & 10 & Not utilized\\
UMass (non-test subset) & 46 & Not utilized\\
\hline
Total & 155 & Unlabeled pool \\
\hline
\end{tabular}
\end{table}

\begin{figure}[!th]
\centering
\includegraphics[width=1 \linewidth]{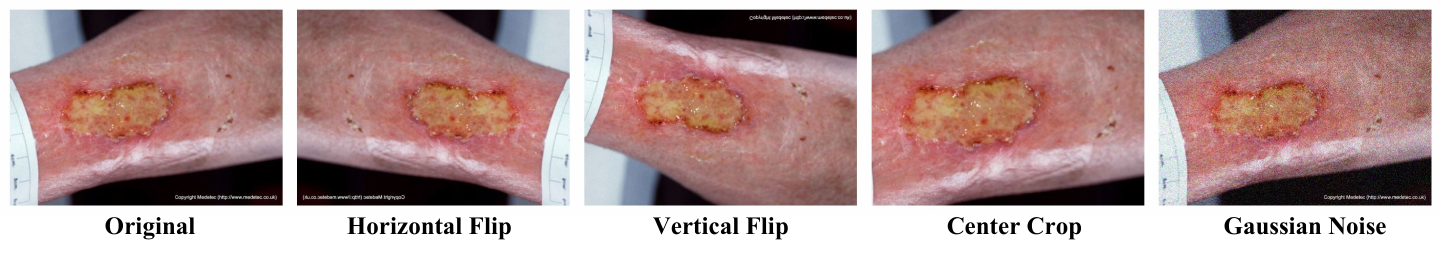}
\caption{Example wound image augmentations used to increase appearance diversity while preserving clinical content. From left to right: original image, horizontal flip, vertical flip, center crop (scale = 0.8), and Gaussian noise (mean = 0, $\sigma = 30$).}
\label{fig:aug_wound}
\end{figure}

To increase appearance diversity while preserving the underlying clinical content, we applied a set of image augmentation operations to each wound image including horizontal flipping, vertical flipping, center cropping, and additive Gaussian noise. Figure~\ref{fig:aug_wound} shows an example of each augmentation applied to the same wound image. 
Together with the original images, these augmented views expanded the dataset from 155 to 620 images in total, which were then used for teacher CoT generation.

\begin{figure}[!h]
\centering
\begin{tcolorbox}[
  colback=gray!5,
  colframe=gray!40,
  boxrule=0.5pt,
  arc=4pt,
  left=6pt, right=6pt, top=5pt, bottom=5pt,
  title={\small \textbf{Teacher-Generated CoT Example}},
  fonttitle=\small,
  coltitle=black,
]
\begin{minipage}[t]{0.30\linewidth}
\vspace{0pt}
\includegraphics[width=\linewidth]{\detokenize{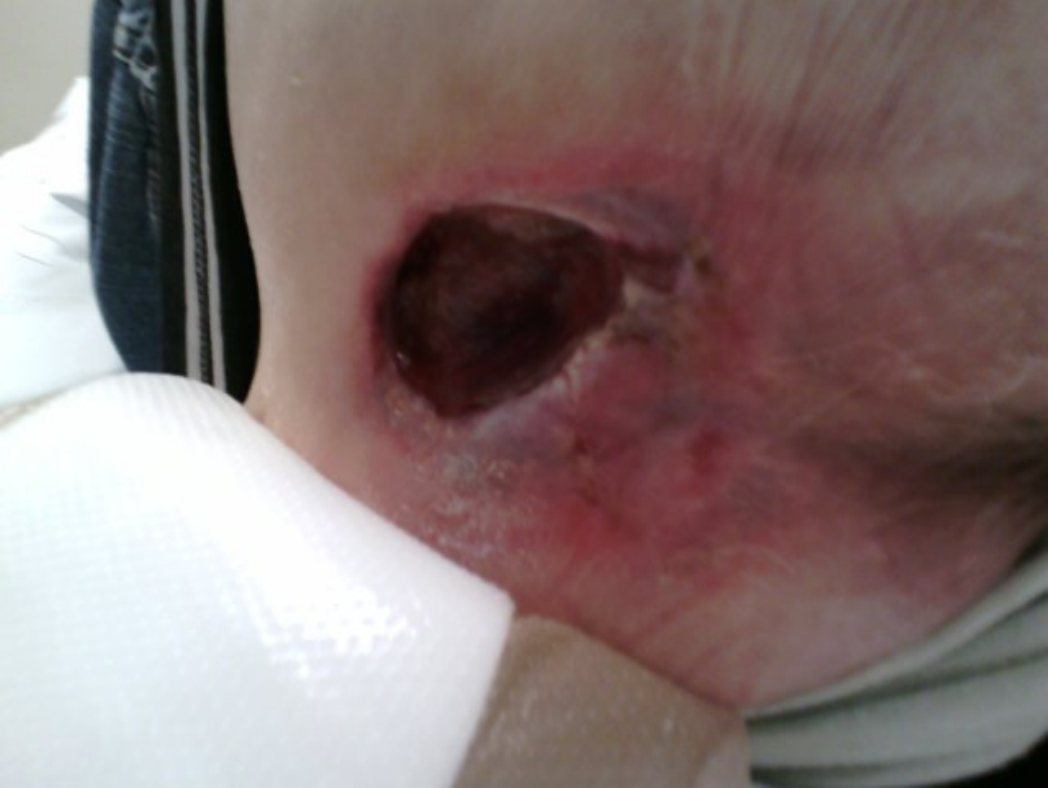}}
\end{minipage}\hfill
\begin{minipage}[t]{0.67\linewidth}
\vspace{8pt}
\small
\textbf{GPT-5.1 Output}\\[4pt]
\texttt{\textless think\textgreater} Redness and irritation are visible around the wound edges, but there is no clear purulent drainage, overt swelling, or obvious spreading cellulitis pattern that strongly suggests infection based solely on this image. The central wound bed appears dark but not clearly indicative of pus. With only visual data, it is not definitively infected. \texttt{\textless /think\textgreater}\\[3pt]
\texttt{\textless answer\textgreater A\textless /answer\textgreater}
\end{minipage}
\end{tcolorbox}
\caption{Example Stage 1 pair: wound image (left) and teacher-generated rationale with final answer (right).}
\label{fig:teacher_cot_example}
\end{figure}

We then applied the CoT prompt in Figure~\ref{fig:reasoning_prompt} to the teacher model (GPT-5.1) to generate rationale-plus-answer outputs for each image, producing the teacher-generated CoT distillation dataset used in Stage 1 (Section~\ref{sec:stage1}). We denote this dataset as $\mathcal{D}_{\text{distill}} = \{(x_i, r_i, \hat{y}_i)\}_{i=1}^{N}$, where $x_i$ is a wound image, $r_i$ is the teacher-generated rationale, and $\hat{y}_i \in \{\texttt{A},\texttt{B}\}$ is the teacher-generated final answer; in our setup, $N=620$. This design introduces some heterogeneity in wound appearance so the student model learns generalizable reasoning patterns rather than source-specific shortcuts or wound-type-specific shortcuts. An example of a teacher-generated rationale is shown in Figure~\ref{fig:teacher_cot_example}.

\subsection{Labeled Infection Dataset for RL Post-Training}

\begin{figure}[!th]
\centering
\includegraphics[width=1 \linewidth]{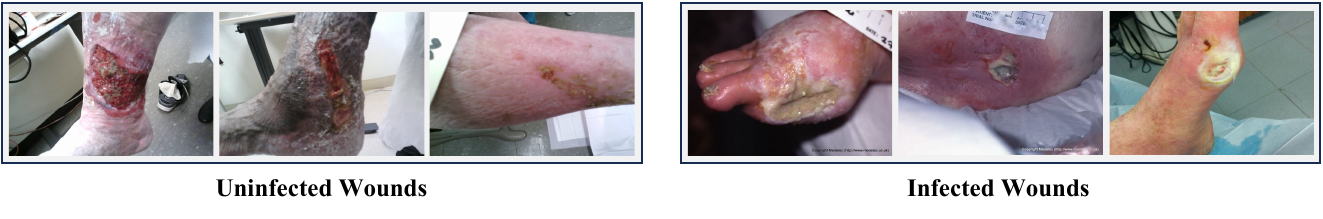}
\caption{Examples of labeled wound images used for RL post-training.}
\label{fig:train_wounds}
\end{figure}

Stage 2 was conducted using a small balanced labeled set of unique wound images. 
We first constructed an original labeled training set of 22 wound images, consisting of 11 infected and 11 uninfected cases. 
The infected images were curated from two clinical sources---the Wound Assessment report~\cite{WUWHS2016Triangle} and \textit{Surgical Dressings and Wound Management}~\cite{MedetecWebsite}---selected because they contain photographically clear examples of infected wounds with prominent visual signs such as purulence, perilesional erythema, and slough (see examples in Figure~\ref{fig:train_wounds}). The 11 uninfected images were drawn from the remaining non-test UMass collection. No image in the Stage 2 training set overlaps with the held-out test set.

To improve robustness under such limited supervision, we applied a set of image augmentations to each image, including horizontal flip, vertical flip, center crop, Gaussian noise, rotation, and shear. By combining these augmentations, we expanded the Stage 2 RL training set, yielding a total of 440 labeled training images. We denote the resulting augmented RL training set as $\mathcal{D}_{\text{Labeled}} = \{(x_i, y_i)\}_{i=1}^{M}$, where $y_i \in \{\texttt{A}, \texttt{B}\}$ is the verified infection label (\texttt{A} = uninfected, \texttt{B} = infected), and $M = 440$. Table~\ref{tab:stage2_split} summarizes the class distribution of the original Stage 2 set, the augmented RL training set, and the held-out test set used for final evaluation.


\begin{table}[!h]
\centering
\caption{Stage 2 labeled training set, augmented RL training set, and held-out test class distribution.}
\label{tab:stage2_split}
\begin{tabular}{lcc}
\hline
& \textbf{Infected} & \textbf{Uninfected}  \\
\hline
Labeled Training Set & 11 & 11  \\
Training + Augmentation & 220 & 220  \\
Held-out UMass Test Set & 28 & 28  \\
\hline
\end{tabular}
\end{table}


The deliberate mismatch between training sources and the held-out evaluation set constitutes a domain shift that makes the final assessment more reflective of real-world deployment conditions. 
\section{Experiments}
\label{sec:experiments}

\subsection{Implementation}
\label{sec:implementation}

All experiments were implemented in Python 3.10 using PyTorch-based libraries including \texttt{torch} 2.6.0, \texttt{transformers} 5.3.0, and \texttt{trl} 0.24.0. The training procedure follows the two-stage paradigm described in Section~\ref{sec:infection_reasoner}, consisting of (1) supervised reasoning distillation and (2) reinforcement learning post-training.

\textbf{Stage 1 (Reasoning Distillation Fine-Tuning, DFT).}  We initialized the student model retrieved from the Hugging Face model repository 
\texttt{Qwen/Qwen3-VL-4B-Thinking} \footnote{\url{https://huggingface.co/Qwen/Qwen3-VL-4B-Thinking}} and fine-tuned it on the distillation dataset $\mathcal{D}_{\text{distill}}$. Full-parameter supervised fine-tuning was performed using the autoregressive language-modeling objective in Equation~\eqref{eq:sft_loss}, where the model was trained to predict the teacher-generated reasoning and final answer tokens conditioned on the input wound image. Training utilized an NVIDIA H100 (80GB) GPU for 300 optimization steps, with batch size 2, gradient accumulation steps 4, learning rate $2\times10^{-5}$, and the AdamW optimizer.

\textbf{Stage 2 (RL Post-Training with GRPO).} 
Stage 2 was initialized from the Stage 1 distilled checkpoint and optimized on the labeled dataset $\mathcal{D}_{\text{Labeled}}$. Policy updates follow the GRPO objective in Equation~\eqref{eq:grpo_loss}, and reward computation follows the verifiable accuracy reward defined in Equation~\eqref{eq:reward}. 
Training utilized four NVIDIA H100 (80GB) GPUs for a conservative 40 optimization steps, with batch size 1, gradient accumulation steps 4, learning rate $1\times10^{-6}$, and the AdamW optimizer, in order to limit overfitting while still allowing task-specific adaptation on the small labeled RL dataset. We set the GRPO group size $G=16$, KL coefficient $\beta=0.5$, and generation temperature to 0.6 during policy rollout. The Stage 2 implementation is adapted from Visual-RFT~\cite{liu2025visual} source code \footnote{\url{https://github.com/Liuziyu77/Visual-RFT}}.

\subsection{Evaluation Metrics}
\label{sec:metrics}

\subsubsection*{Classification Metrics}

Infection classification performance was evaluated on the held-out test set using the following metrics, where $TP$, $TN$, $FP$, and $FN$ denote true positives, true negatives, false positives, and false negatives, respectively, with infected as the positive class:

\begin{itemize}
  \item \textbf{Accuracy:} $\mathrm{ACC} = \dfrac{TP+TN}{TP+TN+FP+FN}$
  \item \textbf{Sensitivity (SEN):} $\mathrm{SEN} = \dfrac{TP}{TP+FN}$ — proportion of infected wounds correctly identified.
  \item \textbf{Specificity (SPC):} $\mathrm{SPC} = \dfrac{TN}{TN+FP}$ — proportion of uninfected wounds correctly identified.
  \item \textbf{Positive Predictive Value (PPV):} $\mathrm{PPV} = \dfrac{TP}{TP+FP}$
  \item \textbf{Negative Predictive Value (NPV):} $\mathrm{NPV} = \dfrac{TN}{TN+FN}$
  \item \textbf{F1-score:} $F_1 = 2 \cdot \dfrac{\mathrm{PPV} \cdot \mathrm{SEN}}{\mathrm{PPV} + \mathrm{SEN}}$
\end{itemize}

\subsubsection*{Rationale Grounding Metric (MLLM-as-a-Judge)}
\label{sec:judge_metrics}

Our objective is to measure \emph{wound-sign grounding}, not narrative persuasiveness. Specifically, given a wound image and a model-generated rationale in \texttt{<think>}...\texttt{</think>} format, we evaluate whether the wound-sign claims in the rationale are supported by visible image evidence. We utilized four independent MLLM judges (GPT-5.1, Gemini-3.1-Pro, Gemini-3.1-Flash, and Claude-Sonnet-4.6), each following the same rubric and evaluation procedure. Full system and user prompts are provided in the Appendix \ref{app:judge_prompt_templates}. For each judge and test case, evaluation has two conceptual steps:

\textbf{Step A --- Text claim extraction.} From \texttt{<think>} only, each rubric sign is assigned one text-claim label:
\texttt{POS} (explicitly asserted present), \texttt{NEG} (explicitly asserted absent), \texttt{UNC} (uncertain/hedged statement), or \texttt{NOT\_MENTIONED} (not discussed in the rationale).

\textbf{Step B --- Visual verification.} Each explicit claim (\texttt{POS}/\texttt{NEG}) is checked against the image and assigned image evidence \texttt{POS}, \texttt{NEG}, or \texttt{UNC} (cannot be determined from visible evidence).

The wound-sign rubric was developed by adapting clinically relevant, visually assessable wound features from established wound-assessment frameworks and infection guidelines~\cite{WUWHS2016Triangle, iwgdf2023dfi}. Specifically, the rubric includes nine signs: purulence/pus, exudate, swelling/edema, erythema/redness, cellulitis or spreading redness, slough/fibrin, necrotic tissue/eschar, friable granulation tissue, and maceration.

\noindent Each sign $k$ for an image $i$ under a judge $j$ receives:
\begin{equation}
\mathrm{score}_{i,k}^{(j)} =
\begin{cases}
1 & \text{if } \texttt{TEXT\_CLAIM}_{i,k}^{(j)} \in \{\texttt{POS},\texttt{NEG}\} \text{ and matches } \texttt{IMAGE\_EVIDENCE}_{i,k}^{(j)}, \\
0 & \text{if } \texttt{TEXT\_CLAIM}_{i,k}^{(j)} \in \{\texttt{POS},\texttt{NEG}\} \text{ and does not match, or } \texttt{IMAGE\_EVIDENCE}_{i,k}^{(j)}=\texttt{UNC}, \\
\texttt{null} & \text{if } \texttt{TEXT\_CLAIM}_{i,k}^{(j)} \in \{\texttt{UNC},\texttt{NOT\_MENTIONED}\}.
\end{cases}
\label{eq:judge_score}
\end{equation}

\noindent For each image $i$ and judge $j$, we compute:
\begin{equation}
C_i^{(j)} = \sum_k \mathbf{1}[\mathrm{score}_{i,k}^{(j)} \neq \texttt{null}], \quad
S_i^{(j)} = \sum_k \mathbf{1}[\mathrm{score}_{i,k}^{(j)} = 1], \quad
A_i^{(j)} =
\begin{cases}
\dfrac{S_i^{(j)}}{C_i^{(j)}} & \text{if } C_i^{(j)} > 0,\\
0 & \text{if } C_i^{(j)} = 0.
\end{cases}
\label{eq:agreement}
\end{equation}

\noindent where $C_i^{(j)}$ is the coverage count and $A_i^{(j)} \in [0,1]$ is the per-image agreement score for judge $j$. In the experimental analysis, we report judge-specific agreement by summarizing $A_i^{(j)}$ across test images for each judge separately. Coverage is retained as an auxiliary count of evaluatable claims, while agreement is the primary rationale-faithfulness metric.

\subsection{Experimental Settings}
\label{sec:baseline}
To evaluate our Infection-Reasoner, we compared it against a diverse set of baseline
models, including proprietary models, open-source MLLMs from both general and medical domains, and
traditional deep-learning image classifiers. The open-source MLLMs are further categorized by parameter size. Specifically, our evaluation includes the following models:

\begin{itemize}
  \item \textbf{Proprietary Models}: GPT-5.1 (2025-11-13 version) \& GPT-5.2 (2025-12-11 version)~\cite{gpt5}, Gemini-3.1-Flash \& Gemini-3.1-Pro~\cite{gemini3pro}, and Claude-Sonnet-4.6~\cite{claude4.6}.
  \item \textbf{Medical MLLMs}: MedVLM-R1~\cite{pan2025medvlmR1}, MedGemma~\cite{google2025medgemma}, HuatuoGPT-Vision~\cite{zhang2023huatuogpt}, and Lingshu~\cite{xu2025lingshu}.
  \item \textbf{General-purpose MLLMs}: Qwen3-VL-Thinking~\cite{qwen3vl}, Qwen3.5~\cite{qwen3.5}, and Llama-3.2-Vision~\cite{grattafiori2024llama}.
  \item \textbf{Traditional Deep Learning Classifiers}: EfficientNet-B0~\cite{efficientnet}, DeiT-Base~\cite{deit}, and SCARWID~\cite{scarwid}.
\end{itemize}

For MLLM baselines, to encourage stable and comparable reasoning behavior, we evaluated each model on the test set across five independent runs using near-deterministic decoding. Specifically, we set the sampling temperature to 0 whenever supported; for Qwen variants, whose implementation does not permit an exact zero value, we utilized $1\times10^{-6}$ as a practical near-zero approximation. For DL classifiers,  we trained them using augmented images from teacher-annotated wound image set combined with the labeled infection dataset.

\subsection{Model Classification Evaluation}
\label{sec:classification_eval}

\begin{table*}[!ht]
\centering
\caption{
Performance comparison of Infection-Reasoner against baseline models on the wound infection test set. 
For MLLMs, each model is evaluated over five independent runs (temperature $=0$ when supported; $1\times10^{-6}$ for Qwen variants), and values are reported as mean $\pm$ standard deviation. For traditional deep classifiers, we report single deterministic point estimates. 
}
\label{tab:main_results}
\resizebox{\textwidth}{!}{
\begin{tabular}{lcccccc}
\hline
\textbf{Method} & \textbf{ACC} & \textbf{SEN} & \textbf{SPC} & \textbf{PPV} & \textbf{NPV} & \textbf{F1} \\
\hline

\multicolumn{7}{c}{\textit{Proprietary Models}} \\
\hline
Claude-Sonnet-4.6 & 56.4 $\pm$ 3.1 & 100 $\pm$ 0.0 & 12.9 $\pm$ 6.2 & 53.5 $\pm$ 1.8 & 100 $\pm$ 0.0 & 69.7 $\pm$ 1.5 \\
Gemini-3-Flash & 72.1 $\pm$ 1.8 & 86.4 $\pm$ 2.7 & 57.9 $\pm$ 3.5 & 67.3 $\pm$ 1.8 & 81.1 $\pm$ 2.9 & 75.6 $\pm$ 1.6 \\
Gemini-3.1-Pro & 75.4 $\pm$ 1.8 & 86.4 $\pm$ 3.5 & 64.3 $\pm$ 0.0 & 70.7 $\pm$ 0.9 & 82.7 $\pm$ 3.7 & 77.8 $\pm$ 1.9 \\
GPT-5.1 (Teacher) & 81.8 $\pm$ 2.1 & 80.0 $\pm$ 1.8 & 83.6 $\pm$ 3.6 & 83.1 $\pm$ 3.2 & 80.7 $\pm$ 1.6 & 81.5 $\pm$ 1.9 \\
GPT-5.2 & 69.6 $\pm$ 3.0 & 59.3 $\pm$ 3.6 & 80.0 $\pm$ 2.9 & 74.8 $\pm$ 3.6 & 66.3 $\pm$ 2.6 & 66.1 $\pm$ 3.5 \\

\hline
\multicolumn{7}{c}{\textit{Open-source Models ($<$ 10B)}} \\
\hline
MedVLM-R1-2B [Medical] & 53.6 $\pm$ 0.0 & 96.4 $\pm$ 0.0 & 10.7 $\pm$ 0.0 & 51.9 $\pm$ 0.0 & 75.0 $\pm$ 0.0 & 67.5 $\pm$ 0.0 \\
MedGemma-4B [Medical] & 50.0 $\pm$ 0.0 & 92.9 $\pm$ 0.0 & 7.1 $\pm$ 0.0 & 50.0 $\pm$ 0.0 & 50.0 $\pm$ 0.0 & 65.0 $\pm$ 0.0 \\
Lingshu-7B [Medical] & 52.1 $\pm$ 1.3 & 97.9 $\pm$ 1.8 & 6.4 $\pm$ 1.4 & 51.1 $\pm$ 0.7 & 76.7 $\pm$ 20.0 & 67.2 $\pm$ 1.0 \\
Qwen3-VL-4B-Thinking & 52.5 $\pm$ 1.8 & 96.4 $\pm$ 0.0 & 8.6 $\pm$ 3.6 & 51.4 $\pm$ 1.0 & 67.7 $\pm$ 10.0 & 67.0 $\pm$ 0.9 \\
Qwen3.5-4B & 60.7 $\pm$ 0.0 & 100 $\pm$ 0.0 & 21.4 $\pm$ 0.0 & 56.0 $\pm$ 0.0 & 100.0 $\pm$ 0.0 & 71.8 $\pm$ 0.0 \\

\hline
\multicolumn{7}{c}{\textit{Open-source Models ($>$ 10B)}} \\
\hline
Lingshu-32B [Medical] & 56.4 $\pm$ 1.8 & 95.7 $\pm$ 1.4 & 17.1 $\pm$ 2.7 & 53.6 $\pm$ 1.1 & 79.8 $\pm$ 6.8 & 68.7 $\pm$ 1.2 \\
HuatuoGPT-V-34B [Medical] & 55.7 $\pm$ 1.3 & 100 $\pm$ 0.0 & 11.4 $\pm$ 2.7 & 53.0 $\pm$ 0.8 & 100 $\pm$ 0.0 & 69.3 $\pm$ 0.6 \\
Qwen3-VL-32B-Thinking & 60.7 $\pm$ 1.1 & 91.4 $\pm$ 1.8 & 30.0 $\pm$ 1.8 & 56.6 $\pm$ 0.7 & 77.9 $\pm$ 3.4 & 69.9 $\pm$ 0.9 \\
Qwen3.5-27B & 67.1 $\pm$ 0.9 & 85.7 $\pm$ 0.0 & 48.6 $\pm$ 1.8 & 62.5 $\pm$ 0.8 & 77.3 $\pm$ 0.6 & 72.3 $\pm$ 0.5 \\
Llama-3.2-90B-Vision-Instruct & 61.1 $\pm$ 1.3 & 96.4 $\pm$ 0.0 & 25.7 $\pm$ 2.7 & 56.5 $\pm$ 0.9 & 87.7 $\pm$ 1.2 & 71.3 $\pm$ 0.7 \\

\hline
\multicolumn{7}{c}{\textit{Traditional Deep Learning Classifiers}} \\
\hline
EfficientNet-B0 & 62.5 & 46.4 & 78.6 & 68.4 & 59.5 & 55.3 \\
DeiT-Base & 62.5 & 53.6 & 71.4 & 65.2 & 60.6 & 58.8 \\
SCARWID (Image-Text input) $^a$ & 76.8 & 75.0 & 78.6 & 77.8 & 75.9 & 76.3 \\

\hline
\multicolumn{7}{c}{\textit{Fine-tuned Models}} \\
\hline
Infection-Reasoner-4B [DFT Only] & 77.1 $\pm$ 1.3 & 73.6 $\pm$ 1.8 & 80.7 $\pm$ 2.9 & 79.3 $\pm$ 2.4 & 75.3 $\pm$ 1.1 & 76.3 $\pm$ 1.2 \\
Infection-Reasoner-4B [GRPO Only] & 74.3 $\pm$ 2.4 & 88.6 $\pm$ 1.4 & 60.0 $\pm$ 4.2 & 69.0 $\pm$ 2.4 & 84.0 $\pm$ 2.2 & 77.5 $\pm$ 1.8 \\
\textbf{Infection-Reasoner-4B [DFT + GRPO]} & \textbf{86.8 $\pm$ 0.9} & \textbf{86.4 $\pm$ 1.4} & \textbf{87.1 $\pm$ 1.8} & \textbf{87.1 $\pm$ 1.5} & \textbf{86.6 $\pm$ 1.2} & \textbf{88.1 $\pm$ 1.0} \\
\hline
\end{tabular}
}

\caption*{%
  \footnotesize
  \textit{%
   $^a$ SCARWID uses GPT-4o-generated wound captions as textual inputs.\\
   $^*$ Throughout the paper we refer to Infection-Reasoner-4B [DFT + GRPO] simply as Infection-Reasoner.\\
  }%
}

\end{table*}


Table~\ref{tab:main_results} presents classification results across all evaluated models. Among all evaluated models, \textbf{Infection-Reasoner-4B [DFT+GRPO]} achieves the strongest and most balanced performance, reaching ACC=$86.8 \pm 0.9$, SEN=$86.4 \pm 1.4$, SPC=$87.1 \pm 1.8$, and NPV=$86.6 \pm 1.2$. 
The balance in sensitivity and specificity is clinically meaningful, since models that maximize infection recall at the expense of specificity may over-flag non-infected wounds, while models that favor specificity too strongly risk missing true infections.

Infection-Reasoner outperforms all proprietary foundation model baselines evaluated in CoT-prompted zero-shot settings, achieving the best overall calibration. Relative to GPT-5.1, which serves as both the teacher model and the strongest proprietary baseline in terms of overall balance, Infection-Reasoner improves ACC by \textbf{+5.0} points ($86.8$ vs.\ $81.8$) and SEN by \textbf{+6.4} points ($86.4$ vs.\ $80.0$), while also slightly improving SPC ($87.1$ vs.\ $83.6$). 
Gemini-3.1-Pro and Gemini-3-Flash both achieve similarly high sensitivity ($86.4$), indicating strong ability to detect infected wounds, but this comes at the cost of much lower specificity ($64.3$ and $57.9$, respectively). Our Infection-Reasoner matches the sensitivity of Gemini-3.1-Pro while improving specificity by \textbf{+22.8} points. 
Claude-Sonnet-4.6 represents the most extreme class-bias behavior, achieving SEN=$100$ but collapsing to SPC=$12.9$, effectively flagging nearly all wounds as infected and providing little practical triage discrimination. Taken together, these results suggest that while several proprietary MLLMs can achieve strong infection sensitivity, Infection-Reasoner is substantially better calibrated for the sensitivity--specificity trade-off required in wound infection assessment, while also showing lower run-to-run standard deviations and therefore suggesting better consistency and more reliable point-of-care behavior.

Statistical significance testing further supports these comparisons. Using a two-proportion z-test on the held-out test set with significance defined at $\alpha = 0.05$, Infection-Reasoner significantly outperformed GPT-5.2 ($p=0.0020$), Gemini-3-Flash ($p=0.0176$), Gemini-3.1-Pro ($p=0.0285$), and Claude-Sonnet-4.6 ($p=0.0001$). Although Infection-Reasoner also achieved higher observed accuracy than GPT-5.1, this difference did not reach statistical significance ($p=0.1040$), likely in part due to the limited evaluation sample size. Nevertheless, this comparison remains practically important, as Infection-Reasoner uses only 4B parameters while achieving performance comparable to, and numerically better than the proprietary frontier model like GPT-5.1. 

A similar pattern appears across open-source MLLMs at both small ($<10$B) and large ($>10$B) scales: many achieve very high sensitivity but low specificity, suggesting a strong bias toward predicting the positive infection class. This behavior is especially pronounced among smaller medical VLMs. MedVLM-R1-2B achieves SEN=$96.4$ but SPC=$10.7$; Lingshu-7B reaches SEN=$97.9$ but SPC=$6.4$; and MedGemma-4B attains SEN=$92.9$ but SPC=$7.1$. 
Although larger open-source models show somewhat improved specificity, they still remain substantially biased toward positive infection predictions.

Traditional deep vision classifiers exhibit the opposite tendency. EfficientNet-B0 and DeiT-Base both achieve ACC=$62.5$, but with relatively low sensitivity ($46.4$ and $53.6$, respectively), indicating that they miss many infected wounds. This is consistent with models trained on both pseudo-labels generated by GPT-5.1 and ground-truth labeled datasets, which appear to favor the uninfected class under traditional supervised learning. SCARWID, a multimodal wound infection classifier that combines wound images with GPT-generated wound descriptions, remained the best-performing classical baseline (ACC=$76.8$, SEN=$75.0$, SPC=$78.6$). Nevertheless, it still underperforms Infection-Reasoner in overall accuracy and clinical utility, as it does not generate an explicit medical rationale.

\begin{figure*}[!ht]
\centering
\begin{tcolorbox}[
  colback=gray!5,
  colframe=gray!40,
  boxrule=0.5pt,
  arc=4pt,
  left=6pt, right=6pt, top=5pt, bottom=5pt,
  title={\small \textbf{Example of VLM Reasoning Outputs on the Same Wound Image}},
  fonttitle=\small,
  coltitle=black,
]
\begin{minipage}[t]{0.24\textwidth}
\vspace{0pt}
\centering
\vspace*{122pt}
\includegraphics[width=\linewidth]{\detokenize{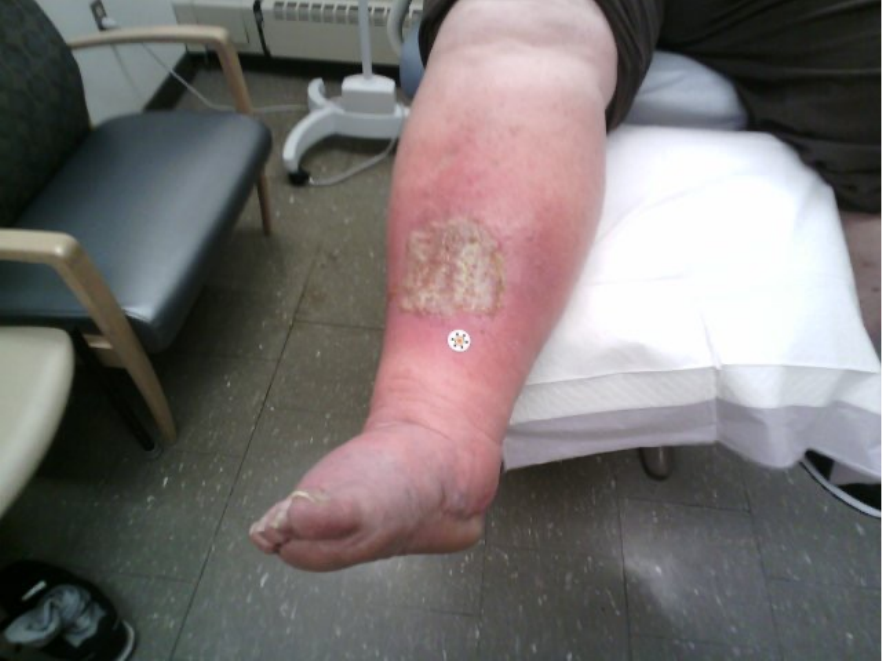}}\\[30pt]
\raggedright\small
\textbf{Question:} Based on visual appearance, is the wound in this image infected?\\[3pt]
\textbf{Choice:} (A) No, (B) Yes.\\[3pt]
\textbf{Prompt setting:} CoT prompt from Fig.~\ref{fig:reasoning_prompt}.\\[3pt]
\textbf{Ground truth:} (B)
\end{minipage}\hfill
\begin{minipage}[t]{0.75\textwidth}
\vspace{0pt}
\small

\begin{tcolorbox}[colback=white,colframe=black!60,boxrule=0.35pt,arc=2pt,left=4pt,right=4pt,top=3pt,bottom=3pt,title=\textbf{GPT-5.1}]
\texttt{\textless think\textgreater} The image shows a lower leg with \textcolor{green!50!black}{significant erythema} extending beyond the wound margins, \textcolor{green!50!black}{diffuse swelling}, and \textcolor{red!80!black}{warmth is implied by the redness}. The wound bed appears covered with \textcolor{green!50!black}{yellowish slough} and there is surrounding skin discoloration. The extent of \textcolor{green!50!black}{redness and edema} beyond the wound edges suggests local infection or \textcolor{green!50!black}{cellulitis} rather than simple inflammation from a healing wound. Based solely on these visual signs, infection is more likely than not.\texttt{\textless /think\textgreater}\\
\texttt{\textless answer\textgreater \textbf{B}\textless /answer\textgreater}
\end{tcolorbox}

\begin{tcolorbox}[colback=white,colframe=black!60,boxrule=0.35pt,arc=2pt,left=4pt,right=4pt,top=3pt,bottom=3pt,title=\textbf{Infection-Reasoner [DFT Only]}]
\texttt{\textless think\textgreater} The wound on the lower leg appears to have a central area of \textcolor{green!50!black}{yellowish slough} or \textcolor{red!80!black}{necrotic tissue}, but the surrounding skin \textcolor{red!80!black}{does not show clear signs of acute infection such as intense erythema, marked swelling}, purulent drainage, or obvious abscess formation. \textcolor{red!80!black}{The redness is somewhat diffuse but not clearly localized or intensely inflamed}. Based solely on visual cues and without systemic/clinical data, it is more consistent with a non-infected chronic wound with slough rather than an overtly infected one. \texttt{\textless /think\textgreater}\\
\texttt{\textless answer\textgreater \textbf{A}\textless /answer\textgreater}
\end{tcolorbox}

\begin{tcolorbox}[colback=white,colframe=black!60,boxrule=0.35pt,arc=2pt,left=4pt,right=4pt,top=3pt,bottom=3pt,title=\textbf{Infection-Reasoner [GRPO Only]}]
\texttt{\textless think\textgreater} So, let's look at the image. The wound on the leg has a red area around it, which might indicate inflammation. But wait, the question is about infection. \textit{Infections often show signs like pus, increased redness, warmth, swelling, or odor.} However, the image shows a red area, but is there pus? \textcolor{red!80!black}{The wound itself looks like a scab or a healing area.} \textcolor{orange!90!black}{The redness could be from inflammation}, but maybe not infection. Wait, the options are A (No) or B (Yes). Let's check the visual cues. \textcolor{green!50!black}{The wound doesn't have obvious pus}, and \textcolor{orange!90!black}{the redness might be due to other factors like trauma or irritation}. Infections typically have more severe signs. $...$ \texttt{\textless /think\textgreater}\\
\texttt{\textless answer\textgreater \textbf{A}\textless /answer\textgreater}
\end{tcolorbox}

\begin{tcolorbox}[colback=white,colframe=black!60,boxrule=0.35pt,arc=2pt,left=4pt,right=4pt,top=3pt,bottom=3pt,title=\textbf{Infection-Reasoner [DFT+GRPO]}]
\texttt{\textless think\textgreater} The wound on the lower leg appears to have a central area of \textcolor{green!50!black}{yellowish slough} or \textcolor{red!80!black}{eschar} with \textcolor{green!50!black}{surrounding erythema}. \textcolor{green!50!black}{The redness extends beyond the immediate wound margins and looks fairly intense, suggesting local inflammation} or infection. There is also some \textcolor{green!50!black}{swelling apparent}. Based solely on these visual signs—\textcolor{green!50!black}{erythema}, \textcolor{orange!90!black}{possible edema}, and \textcolor{green!50!black}{slough}—the wound is more consistent with an infected appearance than a clean, non-infected healing wound. \texttt{\textless /think\textgreater}\\
\texttt{\textless answer\textgreater \textbf{B}\textless /answer\textgreater}
\end{tcolorbox}

\end{minipage}
\end{tcolorbox}
\caption{Qualitative rationale comparison across models on the same wound image. \textcolor{green!50!black}{Green} text marks claims consistent with directly observable wound signs. \textcolor{red!80!black}{Red} text marks incorrect reasoning or claims not reliably inferable from the image alone. \textcolor{orange!90!black}{Orange} text marks uncertain/hedged wording (e.g., ``possible'', ``often'').}
\label{fig:model_reasoning_comparison}
\end{figure*}

\subsubsection{Ablation Study to Demonstrate Importance of Individual Training Stages}
The fine-tuned ablation block in Table~\ref{tab:main_results} provides direct empirical evidence of the contribution of each training stage. DFT-only training improves specificity (SPC=$80.7$) but achieves substantially lower sensitivity (SEN=$73.6$), indicating that reasoning distillation alone induces a more conservative decision pattern that tends to under-diagnose infection, which is similar to the teacher's behavior. Although the DFT-only variant improves substantially over the base model (Qwen3-VL-4B-Thinking), it still underperforms GPT-5.1 in overall accuracy (ACC=$77.1$ vs.\ $81.8$), suggesting that teacher-generated reasoning provides useful structural initialization but is not sufficient by itself to maximize downstream wound infection classification performance. In contrast, GRPO-only training yields high sensitivity (SEN=$88.6$) but weak specificity (SPC=$60.0$), indicating that RL post-training without prior wound-specific distillation tends to over-predict the infected class under sparse reward signals. This same tendency is already evident in the base model, which predicts infection for most wounds.

Combining both stages yields a clear complementary effect. The full DFT+GRPO model increases specificity to $87.1\%$ while preserving high sensitivity at $86.4\%$. 
Relative to the single-stage variants, the full model improves accuracy by \textbf{+12.5} points over GRPO-only and by \textbf{+9.7} points over DFT-only. 

\subsubsection{Qualitative Reasoning Analysis of Fine-tuned Models}
\label{sec:qualitative}

Figure~\ref{fig:model_reasoning_comparison} provides a qualitative illustration of the distinct reasoning behaviors produced by each training configuration on the same test wound image.

The teacher model (GPT-5.1) produces a well-structured and clinically grounded rationale, correctly identifying erythema, swelling, yellowish slough, and redness extending beyond the wound margins, which are all  visual indicators of local infection or cellulitis. Its reasoning is systematic and appropriately anchored to visible evidence, leading to the correct answer. However, the output itself still includes the term \textit{warmth}, which cannot be directly inferred from the image alone.

The DFT-only model exhibits a clear qualitative failure mode. Although its output is human-readable and structurally coherent, it contains hallucinated content, such as \textit{necrotic tissue}, which is not clearly observable in the image. Furthermore, the model incorrectly states the absence of infection signs that are visibly present. It describes the surrounding redness as \textit{``not clearly localized or intensely inflamed''} and concludes that the wound is \textit{``more consistent with a non-infected chronic wound with slough''}, despite the prominent erythema and edema visible in the image. This is a visual grounding failure where the model has learned to imitate the teacher's reasoning style without acquiring sufficient reasoning capability to reach the correct conclusion.

The GRPO-only model exhibits a markedly different reasoning style, revealing the limitations of reinforcement learning without cold-start initialization. Rather than systematically evaluating wound signs, the model produces an unfocused, back-and-forth internal dialogue: \textit{``But wait, the question is about infection\ldots Let's check the visual cues\ldots Wait, the options are A or B\ldots''}. This recursive self-questioning resembles the behavior of the default Qwen3-VL-Thinking model in the absence of structured reasoning priors. This behavior likely reflects the absence of adequated reasoning priors, causing the model to explore possible outputs without systematically anchoring its reasoning in visual evidence from the image~\cite{yue2025does}. The final answer (A = uninfected) is incorrect, and the accompanying rationale provides no reliable clinical justification. This example directly illustrates why Infection-Reasoner's training Stage 1 is necessary: GRPO alone is insufficient to impose the structured, wound-sign-aware reasoning output format required for interpretable clinical decision support.

The full two-stage model combines the structural coherence induced by DFT with the label-correcting pressure introduced by GRPO, producing a rationale that is both well-formed and visually grounded. It correctly identifies yellowish slough, surrounding erythema extending beyond the wound margins, and apparent swelling, and integrates these cues into a correct infection prediction. Compared with the DFT-only model, it is more decisive in emphasizing the intensity and spatial extent of the redness, which appears to be the key visual evidence underweighted by the DFT-only variant. Compared with the GRPO-only model, its reasoning is structured and wound-sign-aware rather than self-referential. However, the model may still produce a minor hallucination, namely \textit{eschar} (i.e., a thick, dry, black or brown layer of dead tissue), which is not clearly supported by the image. Overall, this qualitative progression across ablations is consistent with the quantitative pattern shown in Table~\ref{tab:main_results}: each stage contributes a distinct, useful and complementary capability, and their combination produces the most clinically reliable output. Collectively, the two-stage training pipeline not only improves classification accuracy, but also yields qualitatively better reasoning---more structured than RL alone and more visually faithful than distillation alone.

\subsection{Classification Accuracy by Wound Type}

\begin{figure}[!ht]
\centering
\includegraphics[width=0.70\linewidth]{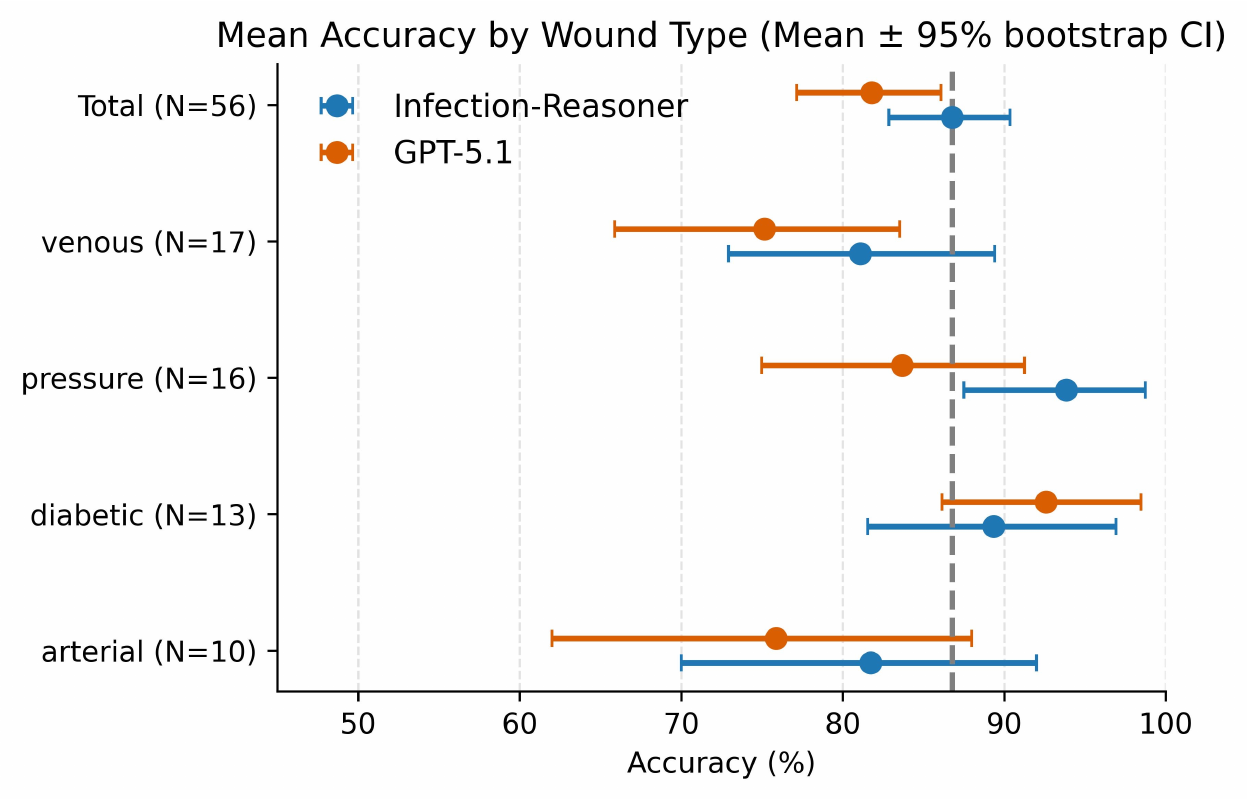}
\caption{Accuracy analysis across wound types. Comparison of classification accuracy between Infection-Reasoner and GPT-5.1 for each wound subgroup in the held-out test set over five runs. Points indicate mean accuracy, and horizontal error bars indicate 95\% bootstrap confidence intervals estimated over test images. The sample size for each subgroup is shown in parentheses.}
\label{fig:accuracy_by_type}
\end{figure}

To assess subgroup behavior beyond aggregate performance, we compared Infection-Reasoner and GPT-5.1 across various wound types in the test set, which included four chronic wound subtypes: venous, arterial, pressure, and diabetic foot ulcers. Figure~\ref{fig:accuracy_by_type} illustrates the mean classification accuracy over five runs for each subgroup, with 95\% bootstrap confidence intervals estimated over test images.

Infection-Reasoner achieved higher mean accuracy than GPT-5.1 on the test set (86.8\% vs.\ 81.8\%). When stratified by wound type, Infection-Reasoner outperformed GPT-5.1 on arterial wounds (81.7\% vs.\ 75.9\%), venous wounds (81.1\% vs.\ 75.1\%), and pressure ulcers (93.8\% vs.\ 83.7\%). The largest subgroup gain was observed for pressure ulcers, where Infection-Reasoner exceeded GPT-5.1 by 10 percentage points. In contrast, GPT-5.1 performed better on diabetic foot ulcers (92.6\% vs.\ 89.4\%), indicating that Infection-Reasoner was not uniformly superior across all wound etiologies.

Several observations emerge from this stratified analysis. First, Infection-Reasoner shows its clearest advantage in pressure, venous, and arterial wounds, suggesting that the benefit of the two-stage training strategy is not limited to the aggregate test set alone. Second, both models perform strongly on diabetic foot ulcers, although GPT-5.1 retains a modest advantage in that subgroup. Third, confidence intervals remain wide for several subgroups, particularly arterial wounds, reflecting the limited subgroup sample sizes. Accordingly, these results should be interpreted as descriptive subgroup evidence rather than definitive proof of wound-type-specific performance differences.

\subsection{Model Reasoning Evaluation}
\label{sec:reasoning_eval}

\subsubsection{MLLM-as-a-Judge Agreement Across Runs}

\begin{figure}[!ht]
\centering
\includegraphics[width=0.68\linewidth]{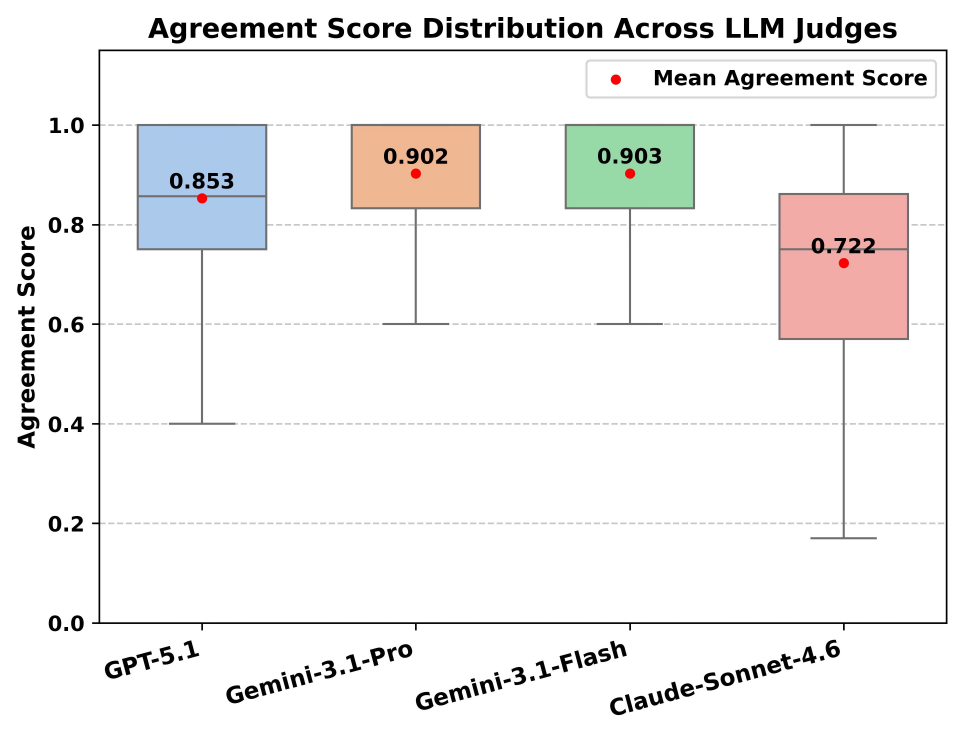}
\caption{Distribution of per-image rationale-grounding agreement scores across four MLLM judges over five runs. For each judge, boxplots summarize the distribution of agreement scores, where each score reflects the fraction of explicitly claimed wound signs in the rationale that are supported by visible image evidence. Red markers denote mean agreement scores.}
\label{fig:judge_agreement_boxplot}
\end{figure}

As described in Section~\ref{sec:judge_metrics}, we evaluate the \emph{visual grounding} of Infection-Reasoner's generated rationales using an MLLM-as-a-Judge framework. Rather than scoring narrative quality or persuasiveness, this metric measures whether wound-sign claims made in the \texttt{<think>} rationale are supported by visible image evidence using a shared clinical rubric. For each image $i$ and judge $j$, we compute a per-image agreement score \(A_i^{(j)} \in [0,1]\) (see Equation~\ref{eq:agreement}), defined as the fraction of explicitly claimed wound signs that are verified by the image among all verifiable claims. Figure~\ref{fig:judge_agreement_boxplot} summarizes the distribution of these agreement scores across test images for four independent MLLM judges.

Gemini-3.1-Flash and Gemini-3.1-Pro achieve the highest mean agreement scores (0.903 and 0.902, respectively), followed by GPT-5.1 (0.853). In contrast, Claude-Sonnet-4.6 yields a lower mean agreement score (0.722) and exhibits greater variance. This pattern appears to reflect a more conservative evaluation style: for some wound signs mentioned by our model, Claude-Sonnet-4.6 marks the evidence as uncertain rather than visually supported, particularly when image quality or local appearance makes assessment difficult. Because our rubric assigns a score of 0 to uncertain judgments for a claimed sign (Equation~\ref{eq:judge_score}), this behavior lowers its overall agreement score and increases score variance. The relatively narrow distributions of the Gemini judges suggest more consistent application of the rubric, whereas the wider spread for Claude-Sonnet-4.6 indicates less stable scoring across cases. GPT-5.1 falls between these two extremes, maintaining relatively high agreement while showing slightly greater variability than the Gemini judges.

Importantly, these agreement scores are not driven by a sparse set of evaluable claims. Across judges and runs, the average coverage is approximately 5 of the 9 rubric-defined wound signs, indicating that Infection-Reasoner produces rationales with a substantial number of explicit, clinically meaningful observations. This is noteworthy because the model was not explicitly trained to enumerate wound signs according to a predefined rubric; instead, this degree of wound-sign coverage emerges from task-oriented reasoning acquired during training. Furthermore, several rubric categories represent closely related clinical appearances. For instance, erythema denotes redness, whereas cellulitis reflects a more extensive pattern of spreading redness. Consequently, in some cases the model may describe visually apparent spreading redness as erythema rather than cellulitis, which is not strictly incorrect but instead reflects partial overlap between related rubric categories. Taken together, the combination of moderate-to-high coverage and high agreement suggests that the model's rationales are not only visually grounded, but also sufficiently specific to support meaningful clinical interpretation.

Overall, the cross-judge pattern suggests that Infection-Reasoner's rationales are generally judged to be well grounded in visible wound evidence by multiple independent evaluators. At the same time, the variation across judges highlights an important limitation of MLLM-based evaluation: agreement estimates depend partly on the evaluator model itself, even when the rubric is fixed. Similar evaluator-dependent variability has also been observed in human wound assessment and image-based clinical decision-making~\cite{wiseman2016inter, nguyen2020machine}. For this reason, we report multiple judges rather than relying on a single evaluator, providing a more reliable assessment of rationale grounding.

\subsubsection{Human Expert Evaluation of Reasoning}

In addition to automated judge-based scoring, we also evaluated rationale quality through manual case review by a a physician wound expert with more than 30 years of clinical experience. To avoid conflating rationale quality with unresolved label disagreement, expert scoring was restricted to the subset of cases for which the expert agreed with the UMass reference label. This yielded an evaluated subset of 34 cases, including 14 infected and 20 uninfected wounds. Within this subset, each rationale was categorized as \textit{Correct}, \textit{Partially Correct}, or \textit{Incorrect} according to the criteria in Table~\ref{tab:rationale_correctness_definition}.

\begin{table*}[!ht]
\centering
\caption{Definition of rationale correctness categories used for expert evaluation.}
\label{tab:rationale_correctness_definition}
\renewcommand{\arraystretch}{1.25}
\setlength{\tabcolsep}{5pt}
\begin{tabular}{p{2.8cm} p{10cm}}
\hline
\textbf{Rating} & \textbf{Definition} \\ 
\hline

\textbf{Correct} 
& The rationale describes clinically plausible visual findings and appropriately supports the predicted infection label. Minor wording issues or omitted details do not alter the clinical interpretation. \\

\textbf{Partially Correct} 
& The rationale supports a reasonable overall conclusion, but includes minor inaccuracies, missing details, overstated findings, or over-reliance on non-specific features as evidence of infection. \\

\textbf{Incorrect} 
& The rationale contains major visual or clinical misinterpretations that make the reasoning inconsistent with the predicted infection label. \\

\hline
\end{tabular}
\end{table*}

\begin{figure}[!ht]
\centering
\includegraphics[width=0.95\linewidth]{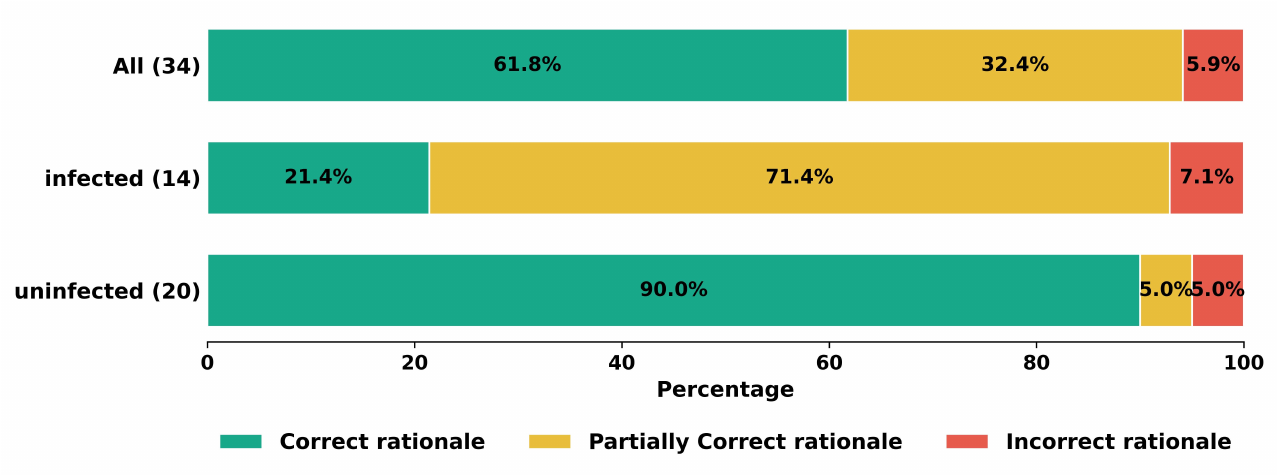}
\caption{Human expert evaluation of Infection-Reasoner rationales on the subset of cases where the wound expert agreed with the UMass reference label. Stacked bars show the percentage of rationales rated as \textit{Correct}, \textit{Partially Correct}, or \textit{Incorrect} for all evaluated cases and for the infected and uninfected strata.}
\label{fig:expert_eval_reasoning}
\end{figure}

Figure~\ref{fig:expert_eval_reasoning} summarizes the distribution of expert ratings. Across all evaluated cases, 61.8\% of rationales were rated as \textit{Correct}, 32.4\% as \textit{Partially Correct}, and 5.9\% as \textit{Incorrect}. Performance was notably better for uninfected wounds, for which 90.0\% of rationales were rated as \textit{Correct}. In contrast, infected wounds were more challenging: only 21.4\% of rationales were rated as \textit{Correct}, whereas 71.4\% were rated as \textit{Partially Correct} and 7.1\% as \textit{Incorrect}.

Expert comments suggest that this gap was driven more by omissions and clinically inaccurate descriptions of infection-related signs than by implausible reasoning. In infected cases, the model often recognized a broadly concerning appearance but sometimes over-relied on features such as slough, necrosis, undermining, or generalized redness, which the reviewing expert considered insufficiently specific, on their own, to support a confident infection-positive interpretation. The expert also noted confusion between infection-related erythema or cellulitis and other causes of redness or discoloration, including ischemia, dependent rubor, Stage~1 pressure injury erythema, and moisture-associated skin damage. In some cases, the rationale omitted important findings such as tunneling, exposed bone, or pigmentary change, or described wound anatomy imprecisely. Overall, Infection-Reasoner usually produced clinically acceptable reasoning, especially for uninfected wounds, while leaving room for improvement in infected cases.


\subsubsection{Qualitative Case Analysis}
\label{sec:qualitative_case_analysis}

To further illustrate these expert-observed reasoning patterns, we present representative qualitative cases spanning expert-aligned correct, partially correct, and incorrect model outputs.

\begin{figure*}[!ht]
\centering
\begin{tcolorbox}[
  colback=gray!5,
  colframe=gray!40,
  boxrule=0.5pt,
  arc=4pt,
  left=6pt, right=6pt, top=5pt, bottom=5pt,
  title={\small \textbf{Example of an Expert-Aligned Correct Case}},
  fonttitle=\small,
  coltitle=black,
]
\begin{minipage}[t]{0.24\textwidth}
\vspace{20pt}
\centering
\includegraphics[width=\linewidth]{\detokenize{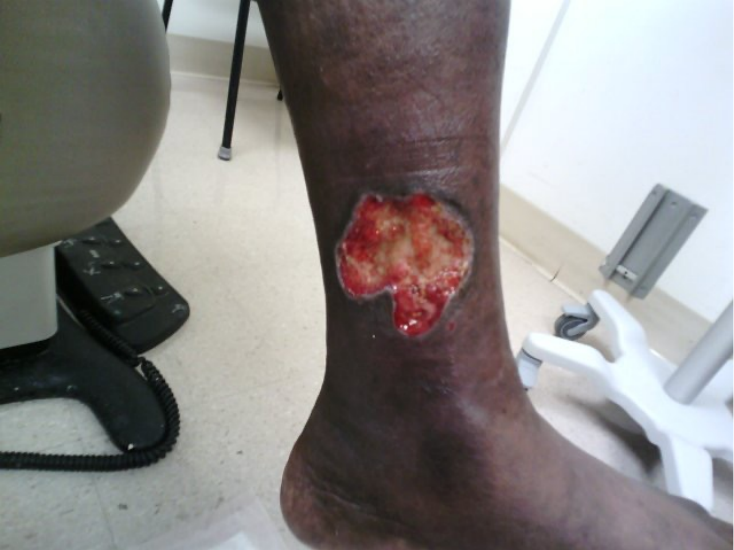}}\\[10pt]
\raggedright\small
\textbf{Ground truth:} (A)
\end{minipage}\hfill
\begin{minipage}[t]{0.75\textwidth}
\vspace{0pt}
\small

\begin{tcolorbox}[colback=white,colframe=black!60,boxrule=0.35pt,arc=2pt,left=4pt,right=4pt,top=3pt,bottom=3pt,title=\textbf{Infection-Reasoner Output}]
\texttt{\textless think\textgreater} The wound shows a red granulating base with some yellowish slough but no obvious purulent exudate, no marked surrounding erythema, edema, or streaking that would clearly indicate infection. The borders appear relatively well-defined without overt cellulitis. Based solely on visual cues, it does not clearly demonstrate classic signs of infection. \texttt{\textless /think\textgreater}\\
\texttt{\textless answer\textgreater \textbf{A}\textless /answer\textgreater}
\end{tcolorbox}

\begin{tcolorbox}[colback=white,colframe=black!60,boxrule=0.35pt,arc=2pt,left=4pt,right=4pt,top=3pt,bottom=3pt,title=\textbf{Expert Evaluation}]
\textbf{Rationale Correctness:} Correct\\[2pt]
\textbf{Comment:} Good description; could add ``rim of hyperpigmentation''.
\end{tcolorbox}

\end{minipage}
\end{tcolorbox}
\caption{Expert-aligned correct-case example. Infection-Reasoner predicts \textbf{A} (uninfected), consistent with the ground truth and expert evaluation.}
\Description{A qualitative correct-case figure with a wound photo and ground-truth label A on the left, and two stacked text boxes on the right: Infection-Reasoner output predicting A and expert evaluation marking the rationale as correct.}
\label{fig:correct_case_expert_aligned}
\end{figure*}

Figure~\ref{fig:correct_case_expert_aligned} shows a representative  expert-aligned correct case. The model correctly predicts the wound as uninfected and supports this decision with clinically plausible negative findings, including the absence of obvious purulence, marked erythema, edema, and overt cellulitis. The expert rated the rationale as \textit{Correct}, noting only a minor omission (rim hyperpigmentation) that does not materially alter the clinical interpretation. 

\begin{figure*}[!ht]
\centering
\begin{tcolorbox}[
  colback=gray!5,
  colframe=gray!40,
  boxrule=0.5pt,
  arc=4pt,
  left=6pt, right=6pt, top=5pt, bottom=5pt,
  title={\small \textbf{Example of a Partially Correct Reasoning Case}},
  fonttitle=\small,
  coltitle=black,
]
\begin{minipage}[t]{0.24\textwidth}
\vspace{20pt}
\centering
\includegraphics[width=\linewidth]{\detokenize{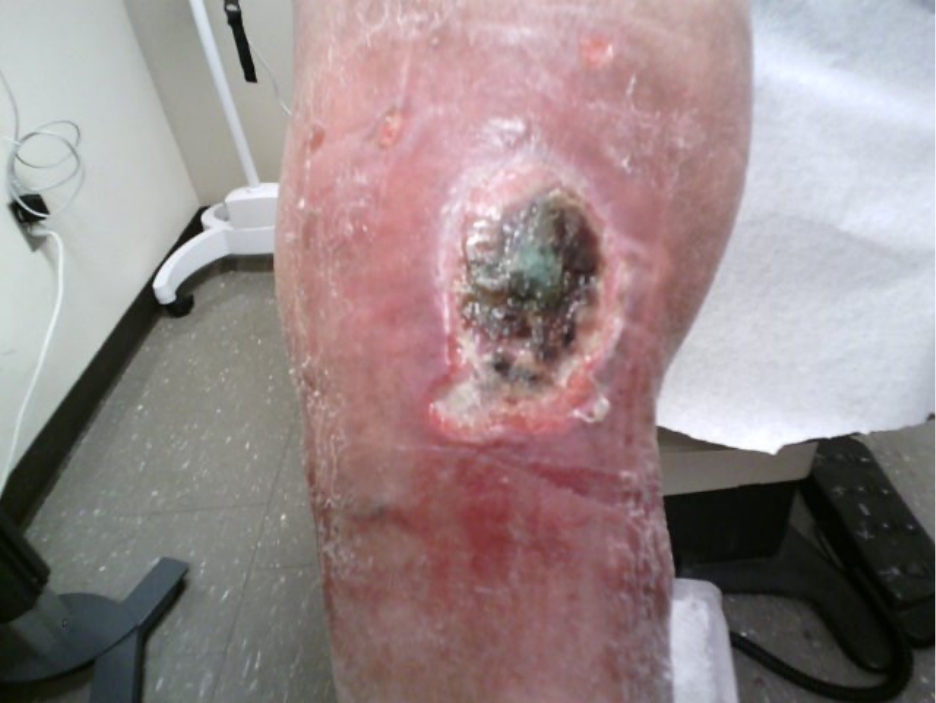}}\\[10pt]
\raggedright\small
\textbf{Ground truth:} (B)
\end{minipage}\hfill
\begin{minipage}[t]{0.75\textwidth}
\vspace{0pt}
\small

\begin{tcolorbox}[colback=white,colframe=black!60,boxrule=0.35pt,arc=2pt,left=4pt,right=4pt,top=3pt,bottom=3pt,title=\textbf{Infection-Reasoner Output}]
\texttt{\textless think\textgreater} The wound shows several visual signs suggestive of infection: there is a large area of necrotic tissue with \textcolor{red!80!black}{greenish slough}, surrounding erythema and edema, and \textcolor{red!80!black}{irregular undermined edges}. The periwound skin appears inflamed and discolored beyond what would be expected from simple irritation, which is consistent with \textcolor{red!80!black}{local infection} rather than just colonization or clean granulation tissue. \texttt{\textless /think\textgreater}\\
\texttt{\textless answer\textgreater \textbf{B}\textless /answer\textgreater}
\end{tcolorbox}

\begin{tcolorbox}[colback=white,colframe=black!60,boxrule=0.35pt,arc=2pt,left=4pt,right=4pt,top=3pt,bottom=3pt,title=\textbf{Expert Evaluation}]
\textbf{Rationale Correctness:} Partially Correct\\[2pt]
\textbf{Comment:} The necrotic center does not appear infected (no exudate, undermining). No greenish slough. The surrounding skin is erythematous consistent with cellulitis (infection) not local infection.
\end{tcolorbox}

\end{minipage}
\end{tcolorbox}
\caption{Example of a partially correct reasoning case. Infection-Reasoner correctly predicts the final label (\textbf{B}: infected), but the expert rates the rationale as \textit{Partially Correct}. \textcolor{red!80!black}{Red} text highlights the specific claims that the wound expert disputed.}
\label{fig:partial_correct_case}
\end{figure*}

\begin{figure*}[!ht]
\centering
\begin{tcolorbox}[
  colback=gray!5,
  colframe=gray!40,
  boxrule=0.5pt,
  arc=4pt,
  left=6pt, right=6pt, top=5pt, bottom=5pt,
  title={\small \textbf{Example of a Failure Case (Completely Incorrect Prediction)}},
  fonttitle=\small,
  coltitle=black,
]
\begin{minipage}[t]{0.24\textwidth}
\vspace{10pt}
\centering
\includegraphics[width=\linewidth]{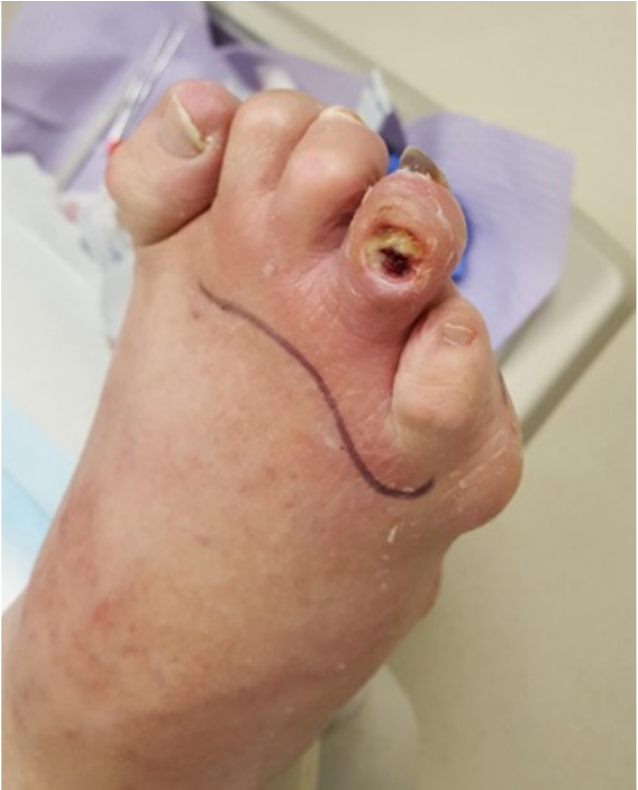}\\[10pt]
\raggedright\small
\textbf{Ground truth:} (B)
\end{minipage}\hfill
\begin{minipage}[t]{0.75\textwidth}
\vspace{0pt}
\small

\begin{tcolorbox}[colback=white,colframe=black!60,boxrule=0.35pt,arc=2pt,left=4pt,right=4pt,top=3pt,bottom=3pt,title=\textbf{Infection-Reasoner Output}]
\texttt{\textless think\textgreater} Toe wound with yellowish slough and central necrotic/eschar tissue, but surrounding skin is only mildly erythematous \textcolor{red!80!black}{without} marked swelling, purulence, or \textcolor{red!80!black}{obvious spreading cellulitis}. No clear visual signs of acute infection such as thick pus, intense redness, \textcolor{red!80!black}{or abscess formation}. Based solely on visual cues, \textcolor{red!80!black}{it appears more like a colonized or healing wound rather than overtly infected}. \texttt{\textless /think\textgreater}\\
\texttt{\textless answer\textgreater \textbf{A}\textless /answer\textgreater}
\end{tcolorbox}

\begin{tcolorbox}[colback=white,colframe=black!60,boxrule=0.35pt,arc=2pt,left=4pt,right=4pt,top=3pt,bottom=3pt,title=\textbf{Expert Evaluation}]
\textbf{Rationale Correctness:} Incorrect\\[2pt]
\textbf{Comment:} The toe is potentially infected with possible underlying bone involvement. The rationale incorrectly dismisses spreading cellulitis (the marked boundary suggests cellulitis tracking), and the absence of abscess cannot be concluded from this single image alone.
\end{tcolorbox}

\end{minipage}
\end{tcolorbox}
\caption{Failure-mode example showing a completely incorrect Infection-Reasoner prediction and rationale. \textcolor{red!80!black}{Red} text marks statements in the model output that the expert disagreed with, judging them as incorrect.}
\label{fig:failure_mode_case}
\end{figure*}

Figure~\ref{fig:partial_correct_case} shows a representative partially correct rationale case. Here, Infection-Reasoner correctly predicts infection and captures that the wound appearance is clinically concerning, suggesting that the model remains useful even when parts of the rationale are incorrect. The expert noted that descriptors such as \textit{greenish slough} and \textit{irregular undermined edges} were not clearly supported by the image, and that the more appropriate interpretation was surrounding erythema consistent with cellulitis which causing infection. The model may have interpreted the central greenish-appearing region as slough, whereas the expert viewed this area more generally as necrotic tissue. Overall, this case suggests that the model's main limitation is not recognizing the need for concern, but precisely characterizing the visual evidence underlying that concern.

Figure~\ref{fig:failure_mode_case} highlights a more severe failure pattern in which both the final prediction and the reasoning are incorrect. Here, the model underestimates infection severity and incorrectly classifies the wound as uninfected. The expert comment indicates that the model wrongly dismisses clinically concerning features, including spreading cellulitis and possible underlying bone exposure. In addition, the rationale makes claims that are not reliably inferable from a single static image, such as the absence of abscess.

\section{Discussion}
\label{sec:Discussion}

\subsection{Summary of Findings}

This study demonstrates that a compact, domain-specialized reasoning VLM can achieve strong performance on chronic wound infection classification from heterogeneous wound photographs while generating clinically interpretable rationales. Across the held-out test set, Infection-Reasoner-4B achieved the best overall diagnostic performance among all evaluated models, reaching 86.8\% accuracy with a balanced trade-off between sensitivity (86.4\%) and specificity (87.1\%).

A key finding is that Infection-Reasoner outperformed GPT-5.1 despite using only 4B parameters. This suggests that, for specialized wound-care imaging tasks, performance may depend more on domain-specific adaptation, structured reasoning supervision, and RL-based alignment to the target task than on model scale alone.

Infection-Reasoner also generalized favorably across multiple wound etiologies. In subgroup analyses, it outperformed GPT-5.1 on arterial, venous, and pressure wounds, with the largest gain observed for pressure ulcers. Although GPT-5.1 retained a modest advantage on diabetic wounds, the overall pattern indicates that the benefits of the proposed two-stage training approach are not limited to a single wound subtype.

Beyond classification performance, Infection-Reasoner produced rationales that were well grounded in visible wound evidence across multiple MLLM evaluators. Taken together, these findings support the potential of compact, fine-tuned reasoning VLMs as effective and interpretable decision-support tools for wound care.

\subsection{Clinician Feedback on Model Reasoning}

The clinician-facing value of Infection-Reasoner depends not only on prediction accuracy, but also on whether its rationales are clinically plausible and anchored to relevant visual evidence. Our expert evaluation involved an important complement to automated MLLM rationale scoring in this regard. On the wound cases where the expert agreed with the reference label, most rationales were rated as either \textit{Correct} or \textit{Partially Correct}, with relatively few rated as outright \textit{Incorrect}. This suggests that Infection-Reasoner typically produces reasoning that is clinically acceptable in broad terms, even when its explanations are not fully expert-level.

A particularly important pattern was the difference between infected and uninfected wounds. For uninfected wounds, the expert rated the large majority of rationales as \textit{Correct}, indicating that the model is relatively reliable at recognizing the absence of overt infection signs when the wound appearance is visually reassuring. In contrast, infected wounds were much more challenging: the dominant error mode was not completely implausible reasoning, but partially correct reasoning that identified some concerning features while failing to prioritize the most clinically discriminative evidence. This distinction is important because it suggests that the model often recognizes that a wound is abnormal, yet still struggles to correctly express the reasoning in the same way an experienced wound expert would.

In conclusion, the clinician feedback strengthens the translational relevance of this work. It suggests that Infection-Reasoner is already capable of producing clinically useful, evidence-aware explanations in many cases, especially for uninfected wounds, while also identifying an important area for future improvement: more precise and expert-aligned reasoning for infection-positive cases.


\subsection{Limitations}

This study has several limitations. First, the labeled dataset used for RL post-training is small, and the held-out test set is also limited in size. Although repeated runs and bootstrap confidence intervals help characterize consistency and variability, some subgroup estimates remain uncertain, particularly for wound-type-specific analyses. As a result, the performance differences reported here should be interpreted as strong preliminary evidence rather than definitive proof of generalization across all wound populations and care settings.

Second, the evaluation is based on wound photographs alone. In real clinical practice, infection assessment is rarely determined from appearance alone; clinicians also consider other contextual information including patient history, pain, odor, drainage characteristics, temperature, wound progression over time, debridement findings and laboratory data. Our model was intentionally designed for image-only decision support from appearance in POC settings, where access to broader diagnostic resources may be limited. Accordingly, Infection-Reasoner should be viewed as a triage-support or decision-support tool rather than a standalone diagnostic system.

Third, although our training data are heterogeneous and include multiple wound etiologies, the final evaluation still reflects a limited sample drawn from a specific study context. Broader external validation across institutions, skin tones, and care workflows is still necessary, which is  is particularly important in wound assessment, where image quality and contextual variability can strongly affect model behavior.

Fourth, rationale evaluation remains an open challenge. The MLLM-as-a-Judge protocol provides a useful way to assess whether explicitly claimed wound signs are visually supported, but the scores depend partly on the specific MLLM selected as the evaluator and the grading rubric provided. Likewise, while the expert review provided valuable clinical insights, it reflects the judgment of a single wound-care expert on a subset of cases. Additional multi-rater expert evaluation would improve reliability and better characterize inter-clinician agreement on rationale quality.

\subsection{Future Work} 

Several research directions could extend this work. The first priority is larger-scale external validation across multiple institutions and wound-care environments. Such evaluation should include much larger dataset size, broader diversity in wound subtype, acquisition device, anatomical site, skin tone, and image quality in order to more rigorously assess robustness and fairness under real-world deployment conditions.

The second direction is the incorporation of clinical context beyond the image itself. Future systems could integrate metadata such as wound duration, dressing history, pain level, healing progression, patient symptoms, or corresponding thermal wound images. Because infection diagnosis often depends on the combination of visual and non-visual evidence, multimodal fusion may improve both classification performance and the specificity of rationales generated.

The last direction involves improving rationale quality, especially for infected wounds. Our results suggest that the next challenge is not only producing visually grounded explanations, but also making those explanations more clinically discriminative and expert-aligned. Instead of a simple CoT prompting, we may use a specific infection diagnosis rubric derived by experts as structural-instruction prompting to control which wound signs the model should consider during the multi-stage finetuning step. 


\section{Conclusion}
\label{sec:conclusion}

We presented Infection-Reasoner, a compact 4B reasoning vision-language model for chronic wound infection classification and evidence-grounded clinical rationale generation from wound photographs. Using a two-stage training strategy that combines teacher-guided reasoning distillation with RL post-training, the model achieved strong performance on a held-out dataset containing four chronic wound types, and produced clinically interpretable rationales linked to visible wound findings. A central result of this study is that while using only 4B parameters, Infection-Reasoner outperformed GPT-5.1 and other large multimodal baselines. This finding highlights an important practical finding for medical AI: in specialized tasks such as wound infection assessment from images, targeted fine-tuning and reasoning-focused post-training can be more important than raw model size alone. At the same time, our expert analysis shows that there remains room for improvement, especially in infection-positive cases where clinically discriminative reasoning is most challenging. Overall, however, the results suggest that Infection-Reasoner is a promising step toward interpretable, compact, and task-specialized multimodal AI for wound decision support at the point of care.

\section*{Acknowledgment}
This work is supported by the National Institutes of Health (NIH) through grant 1R01EB031910-01A1 Smartphone-based wound infection screener by combining thermal images and photographs using deep learning methods. The experiments were performed using computational resources provided by the Academic \& Research Computing group at Worcester Polytechnic Institute.

\section*{Declaration of competing interest}
The authors declare that they have no conflicts of interest.

\bibliographystyle{ACM-Reference-Format}
\bibliography{main}

\section{Appendix}

\subsection{MLLM-as-a-Judge Prompt Templates}
\label{app:judge_prompt_templates}

\begin{tcblisting}{
  colback=gray!5,
  colframe=gray!40,
  boxrule=0.5pt,
  arc=4pt,
  left=6pt, right=6pt, top=5pt, bottom=5pt,
  title={\small \textbf{System Prompt Template}},
  fonttitle=\small,
  coltitle=black,
  listing only,
  breakable,
  listing options={basicstyle=\ttfamily\footnotesize, breaklines=true, columns=fullflexible}
}
You are a meticulous wound-rationale evaluator.

You will be given:
(1) a wound image, and
(2) a model-generated text containing <think>...</think> and <answer>...</answer>.

Your task is to perform TWO steps internally, but return ONE final STRICT JSON only:

Step A: Claim Extraction from <think> only
- Read ONLY the content inside <think>...</think>.
- Extract wound-sign claims for the rubric items below.
- Do NOT use the image when deciding TEXT_CLAIM.
- Do NOT use <answer> for claim extraction unless <think> is missing.
- For each rubric item, assign:
  - POS = the text asserts the sign is present/seen
  - NEG = the text asserts the sign is absent/not seen
  - UNC = the text is uncertain/hedged (e.g., "unclear", "possible", "not definitive", "at risk", "not marked")
  - NOT_MENTIONED = the sign is not discussed
- text_span must be a short exact quote from <think> supporting the label, or "" if NOT_MENTIONED.

Step B: Visual Verification from image only
- Verify whether each explicit TEXT_CLAIM (POS/NEG) is visually supported by the image.
- Use ONLY visible evidence in the image.
- Do NOT infer based on clinical priors, infection patterns, or likely diagnoses.
- Do NOT revise or overwrite TEXT_CLAIM based on the image.
- If the sign cannot be assessed from the image, set IMAGE_EVIDENCE = UNC.

Definitions for IMAGE_EVIDENCE:
- POS = sign is clearly present/visible
- NEG = sign is clearly absent/not visible
- UNC = cannot determine from the image

Scoring:
- SCORE = 1 if TEXT_CLAIM in {POS, NEG} and matches IMAGE_EVIDENCE
- SCORE = 0 if TEXT_CLAIM in {POS, NEG} and does not match IMAGE_EVIDENCE
- SCORE = 0 if TEXT_CLAIM in {POS, NEG} and IMAGE_EVIDENCE = UNC
- SCORE = null if TEXT_CLAIM is UNC or NOT_MENTIONED

Final aggregation:
- coverage_count = number of rubric items with SCORE not null
- supported_count = number of rubric items with SCORE = 1
- agreement_score = supported_count / coverage_count
- If coverage_count = 0, agreement_score = null

Rubric items:
1. purulence_pus
2. exudate
3. swelling_edema
4. erythema_redness
5. cellulitis_spreading_redness
6. slough_fibrin
7. necrotic_tissue_eschar
8. friable_granulation
9. maceration

Important rules:
- Never use the image to change TEXT_CLAIM.
- Never use the text to invent visual evidence.
- Be conservative: prefer UNC over guessing.
- Return STRICT JSON only.
- No markdown, no prose, no code fences.
\end{tcblisting}

\begin{tcblisting}{
  colback=gray!5,
  colframe=gray!40,
  boxrule=0.5pt,
  arc=4pt,
  left=6pt, right=6pt, top=5pt, bottom=5pt,
  title={\small \textbf{User Prompt Template}},
  fonttitle=\small,
  coltitle=black,
  listing only,
  breakable,
  listing options={basicstyle=\ttfamily\footnotesize, breaklines=true, columns=fullflexible}
}
Inputs:
1) Wound image: (attached)
2) Model generation text:
{GENERATION_TEXT}

Return STRICT JSON with this exact schema:
{
  "parsed_think": "string",
  "rubric": {
    "purulence_pus": {
      "TEXT_CLAIM":"POS|NEG|UNC|NOT_MENTIONED",
      "text_span":"string",
      "IMAGE_EVIDENCE":"POS|NEG|UNC",
      "SCORE": 0|1|null,
      "image_rationale":"string"
    },
    "exudate": { ... },
    "swelling_edema": { ... },
    "erythema_redness": { ... },
    "cellulitis_spreading_redness": { ... },
    "slough_fibrin": { ... },
    "necrotic_tissue_eschar": { ... },
    "friable_granulation": { ... },
    "maceration": { ... }
  },
  "coverage_count": 0,
  "supported_count": 0,
  "agreement_score": 0.0
}

Additional instructions:
- parsed_think should contain the extracted content inside <think>...</think>, or "" if absent.
- text_span should be a short exact quote from <think>, or "" if NOT_MENTIONED.
- image_rationale should be one short sentence describing only what is visible, or why it is UNC.
- Keep outputs concise and strictly JSON.
\end{tcblisting}


\end{document}